\documentclass[journal]{IEEEtran}
\usepackage{amsmath,amsfonts}
\usepackage{algorithmic}
\usepackage{algorithm}
\usepackage{array}
\usepackage{textcomp}
\usepackage{stfloats}
\usepackage{verbatim}
\usepackage{graphicx}
\usepackage{cite}
\usepackage{subfigure}
\usepackage{svg}
\usepackage{multirow}
\usepackage{booktabs}
\usepackage{makecell}
\usepackage[colorlinks]{hyperref}

\begin{document}

\title{Semantic-Aware Adversarial Training for Reliable Deep Hashing Retrieval}

\author{Xu~Yuan,~Zheng~Zhang,~\IEEEmembership{Senior~Member,~IEEE,}~Xunguang~Wang,~Lin~Wu
\thanks{This work was supported by Shenzhen Science and Technology Program under Grant RCYX20221008092852077, National Natural Science Foundation of China under Grant 62002085, and Guangdong Basic and Applied Basic Research Foundation under Grant 2023A1515010057. (\textit{Corresponding Author: Zheng Zhang})}
\thanks{X. Yuan and X. Wang are with School of Computer Science and Technology, Harbin Institute of Technology, Shenzhen 518055, China. (E-mail: xuyuan127@gmail.com, xunguangwang@gmail.com)}
\thanks{Z. Zhang is with School of Computer Science and Technology, Harbin Institute of Technology, Shenzhen 518055, China, and also with Peng Cheng Laboratory, Shenzhen 518055, China. (E-mail: darrenzz219@gmail.com)}
\thanks{L. Wu is with Department of Computer Science, Swansea University, UK. (E-mail: lin.wu@uwa.edu.au)}
}

\markboth{IEEE Transactions on Information Forensics and Security, VOL. 18, 2023}%
{Shell \MakeLowercase{\textit{et al.}}: A Sample Article Using IEEEtran.cls for IEEE Journals}

\IEEEpubid{1556-6021~\copyright~2023 IEEE}

\maketitle
\begin{abstract}
Deep hashing has been intensively studied and successfully applied in large-scale image retrieval systems due to its efficiency and effectiveness. Recent studies have recognized that the existence of adversarial examples poses a security threat to deep hashing models, that is, adversarial vulnerability. Notably, it is challenging to efficiently distill reliable semantic representatives for deep hashing to guide adversarial learning, and thereby it hinders the enhancement of adversarial robustness of deep hashing-based retrieval models. Moreover, current researches on adversarial training for deep hashing are hard to be formalized into a unified \textit{minimax} structure. In this paper, we explore Semantic-Aware Adversarial Training (SAAT) for improving the adversarial robustness of deep hashing models. Specifically, we conceive a discriminative mainstay features learning (DMFL) scheme to construct semantic representatives for guiding adversarial learning in deep hashing. Particularly, our DMFL with the strict theoretical guarantee is adaptively optimized in a discriminative learning manner, where both discriminative and semantic properties are jointly considered. Moreover, adversarial examples are fabricated by maximizing the Hamming distance between the hash codes of adversarial samples and mainstay features, the efficacy of which is validated in the adversarial attack trials. Further, we, \textit{for the first time}, formulate the formalized adversarial training of deep hashing into a unified minimax optimization under the guidance of the generated mainstay codes. Extensive experiments on benchmark datasets show superb attack performance against the state-of-the-art algorithms, meanwhile, the proposed adversarial training can effectively eliminate adversarial perturbations for trustworthy deep hashing-based retrieval. Our code is available at \href{https://github.com/xandery-geek/SAAT}{https://github.com/xandery-geek/SAAT}.

\end{abstract}


\section{Introduction}
\IEEEPARstart{W}{ith} the growing of large-scale multimedia data, approximate nearest neighbor (ANN) retrieval \cite{andoni2006near} has received much attention due to its outstanding balance capability of efficiency and effectiveness. Among various ANN methods, hashing \cite{wang2017survey} offers the prominent advantages of mapping high-dimensional data to compact binary codes with low costs in storage and computational complexity. Compared to shallow hashing methods, deep hashing\cite{xia2014supervised,li2016feature,cao2017hashnet,talreja2020deep,yuan2020central,wang2020deep,fan2020deep,hoe2021one,doan2022one,wu2023deep} achieves superior performance by learning a nonlinear hashing function based on deep neural networks in an end-to-end manner.

\begin{figure}[t]
	\begin{center}
		\includegraphics[width=0.48\textwidth]{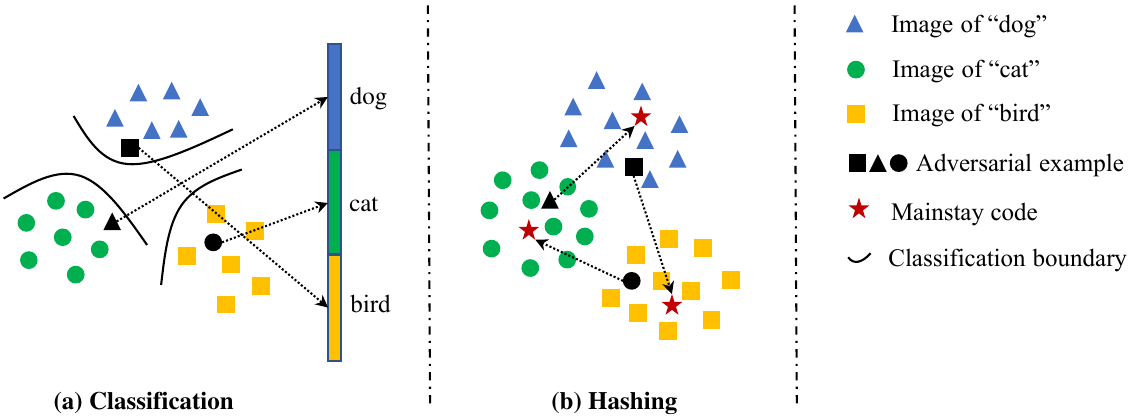}
	\end{center}
	\caption{The comparison between classification and semantic similarity-preserving hashing on adversarial learning.
	In the output space of classification, an adversarial attack with the guidance of labels only needs to make the adversarial samples across the decision boundary. However, an adversarial attack on deep hashing is more challenging because the adversarial samples are expected to blend into the clusters in the embedding space, but there is not an explicit signal to supervise the process. The same problem exists for adversarial training. For classification, adversarial training just maximizes probabilities of adversarial samples on true labels, which is infeasible on hashing due to the lack of explicit supervision signals. Hence, deep hashing needs reliable and discriminative semantic representatives (\textit{i.e.}, mainstay) to represent the image semantics for adversarial attacks and defense.}
	\label{fig:motivation}\vspace{-0.2cm}
\end{figure}

Recent studies \cite{yang2018adversarial,bai2020targeted,wang2021prototype,zhang2021targeted,lu2021smart} have demonstrated that deep hashing-based retrieval models are vulnerable to adversarial examples. Generally, the adversarial examples \cite{szegedy2013intriguing} are crafted by adding imperceptible perturbations to original samples, yet can greatly confuse deep hashing models to retrieve incorrect results. Undoubtedly, such malicious attacks \cite{chen2019generative,zhong2020towards} pose serious security risks to deep hashing-based retrieval systems. For example, in a deep hashing-based face recognition system, adversarial examples can mislead the system and retrieve a non-matching person's face, and thus successfully invade the system. This necessitates studying the adversarial attack on deep hashing models and developing effective defense strategies.

\IEEEpubidadjcol
The current researches working with adversarial samples mostly focus on classification task \cite{li2022decision}, with the exception of a few works \cite{yang2018adversarial,bai2020targeted,wang2021prototype,zhang2021targeted,lu2021smart} studying the adversarial learning on deep hashing-based retrieval. However, these two tasks are quite different in terms of adversarial learning, as shown in Fig. \ref{fig:motivation}. The former learns to classify each sample to its target class under the supervision of class labels. The latter is a similarity-preserving task that maps high-dimensional data to Hamming space and maintains the similarity between the data. It is clear that we can generate adversarial examples that can deceive and out-trick the classification system by minimally recomposing classification boundaries derived from semantic labels. However, for the semantic similarity-preserving hashing-based retrieval task, \textit{there are no explicit representatives (\textit{e.g.}, label) to lead adversarial attacks or defenses}, resulting in more challenges compared to the classification one. Moreover, the \textit{minimax optimization-based adversarial training,} which has shown success in classification, does not seamlessly transfer to deep hashing due to the absence of semantic representations.
Therefore, it is challenging to address the following two problems:
\begin{enumerate}
    \item How could we initialize the optimal and global semantic representatives for adversarial learning in deep semantic-preserving hashing networks?
    \item How could we derive a unified minimax formulation for deep hashing to conduct robust adversarial training?
\end{enumerate}

For the first question, some adversarial attack works \cite{yang2018adversarial,bai2020targeted} heuristically select the hash code of a single sample as each representative for generating its adversarial example, while some others \cite{bai2020targeted, lu2021smart} employ a set of hash codes from multiple relevant samples. Nevertheless, these methods are not reliable and fail to capture globally discriminative semantics, yielding low-quality semantic representatives. Another line of work \cite{wang2021prototype, wang2021targeted} adopts an auxiliary network to learn representative hash codes of label encoding for targeted attacks. Typically, such an auxiliary prototype network is a parameter-sensitive scheme that lacks theoretical guarantees, so the whole adversarial generation is poorly generalized over different datasets and hashing networks. In a nutshell, these works are less efficient to generate high-quality semantic representatives for adversarial deep hashing.

For the second question, there is only one work \cite{wang2021targeted} on adversarial defense for deep hashing that simply reduces the distance between adversarial examples and the benign samples in Hamming space. However, this method is hard to enable robust adversarial training because it is not a standard minimax adversarial optimization problem shown in \cite{madry2017towards}, composed of an \textit{inner maximization} problem and an \textit{outer minimization} problem. 
Notably, such a unified learning framework cannot hold in regular deep hashing-based retrieval networks unless preferable semantic representatives are provided.

\begin{figure*}[t]
	\begin{center}
		\includegraphics[width=0.92\linewidth]{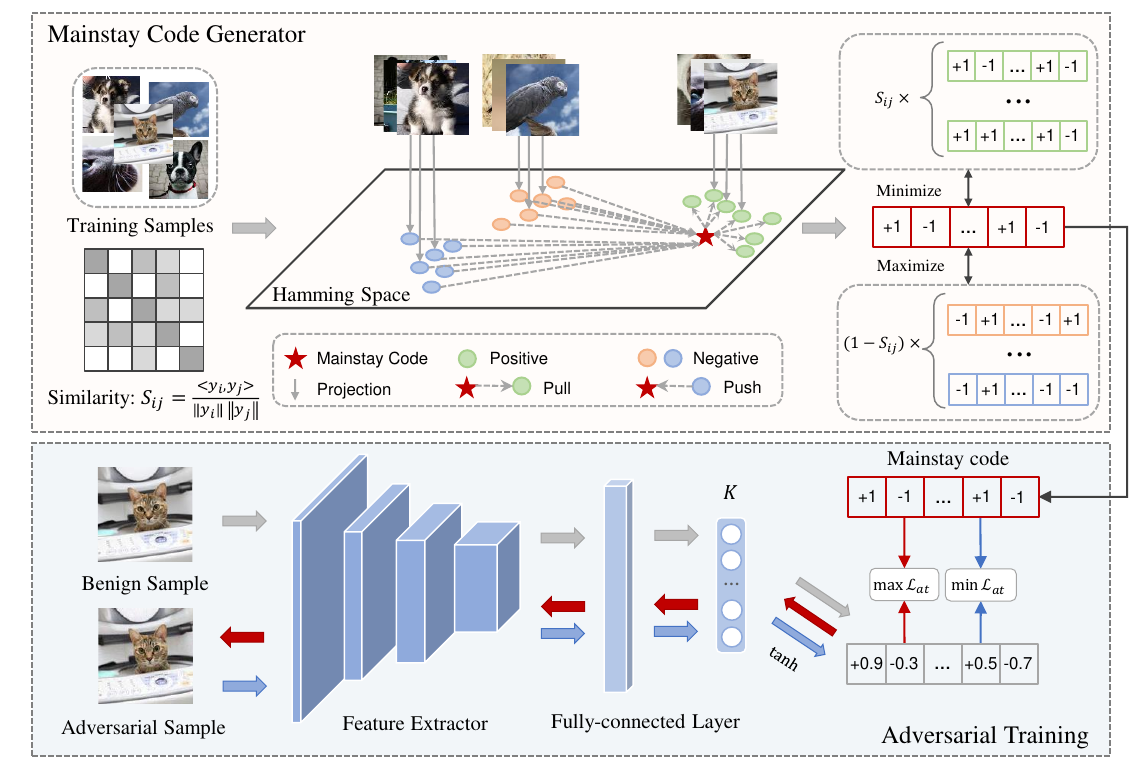}
	\end{center}
    \vspace{-0.2cm}
	\caption{\small The pipeline of the proposed Semantic-Aware Adversarial Training (SAAT) for deep hashing retrieval. The architecture is composed of two mechanisms: the generation branch of representative codes (\textit{i.e.,} mainstay codes) and the minimax-based adversarial training branch.
	In mainstay code generation, all training samples are projected to the Hamming space to form their corresponding hash codes. Then, we build a mainstay code for each class by discriminative learning which pulls the mainstay code closer to the hash code of positives as well as pushes it away from other negatives.
	In adversarial training, we extend adversarial training of deep hashing to a minimax framework \textit{i.e.}, standard adversarial training. As illustrated, the gray, red, and blue arrows indicate forward, backward and forward propagation, respectively. The red arrow means constructing adversarial samples with the supervision of generated mainstay codes and the blue arrow represents inputting adversarial samples for adversarial training. Best viewed in color.}
	\label{fig:framework}
	\vspace{-0.3cm}
\end{figure*}

To overcome the aforementioned issues, this paper constructs discriminative semantic representatives (dubbed \textit{mainstay codes}) for adversarial learning of deep hashing and further proposes Semantic-Aware Adversarial Training (SAAT) with a formalized minimax framework to strengthen the adversarial robustness of deep hashing models. Intuitively, we rethink the characteristics of semantic similarity-preserving hashing-based retrieval tasks. Different from classification, the purpose of retrieval is to return top-n relevant objects instead of one result. Hence, the optimal semantic representative of given sample $\boldsymbol{x}$ in the retrieval task should preserve both similarities with all semantically relevant samples (positives) and dissimilarity with all semantically irrelevant ones (negatives). From this viewpoint, we argue that the semantic representative for adversarial deep hashing is expected to have a minimum Hamming distance from all positives yet a maximum distance from all negatives.

Directly optimizing the above problem is intractable due to the infeasibility of utilizing gradient descent in discrete Hamming space and the costly computational expense incurred by numerous positive and negative samples. 
Owing to the binarization of hash codes, \textit{we allow an efficient approach called Discriminative Mainstay Features Learning (DMFL), which offers rigorous theoretical guarantees and enables the direct acquisition of the optimal solution to this problem}, \textit{i.e.}, the proposed mainstay code.
Then, we transform the adversarial attack (\textit{e.g.}, non-targeted attack) on deep hashing into maximizing the Hamming distance between the hash code of the adversarial example and the mainstay code to efficiently generate the optimal adversarial example.
Furthermore, based on the generated mainstay codes, we formulate the adversarial training on deep hashing as a minimax optimization problem, \textit{i.e.}, a unified and standard adversarial training formula. Under the minimax paradigm, the inner maximization seeks an adversarial example $\boldsymbol{x}^{\prime}$ of a given data $\boldsymbol{x}$ whose hash code maximizes the Hamming distance from the mainstay code, and the outer minimization attempts to find model parameters so that the hash code of $\boldsymbol{x}^{\prime}$ is close to the mainstay code to alleviate the effects caused by the adversarial perturbations.
The overall framework is illustrated in Fig. \ref{fig:framework}. In summary, our main contributions can be summarized below:

\begin{itemize}
    \item We propose a \textbf{S}emantic-\textbf{A}ware \textbf{A}dversarial \textbf{T}raining (\textbf{SAAT}) framework for optimizing reliable deep hashing models. To our best knowledge, \textit{this is the first attempt} to formulate the formalized adversarial training of deep hashing into a unified minimax paradigm guided by well-designed globally optimal semantic representatives.
    \item A discriminative mainstay features learning (DMFL) scheme is well conceived to generate global semantic representatives (named \textit{mainstay codes}) of deep hashing, which can efficiently guide adversarial learning of deep hashing networks. As a consequence, the produced mainstay codes can be simply adaptive to non-targeted and targeted adversarial attacks against deep hashing models.
    \item In the presence of the derived mainstay features, we formalize a well-designed adversarial training strategy under a \textit{minimax} optimization for enhancing the adversarial robustness of deep hashing-based retrieval.
    \item Extensive experiments validate the proposed attack method's superiority over state-of-the-art attacks on deep hashing. Meanwhile, further experiments demonstrate that our SAAT can substantially aid deep hashing networks in resisting multiple adversarial attacks.
\end{itemize}

The remainder of this paper is organized as follows. The related works are reviewed in Section \ref{sec:related_work}. Section \ref{sec:method} describes the proposed semantic-aware adversarial training method, including the problem formulation, the generation of mainstay code, and the attack/defense algorithm. In Section \ref{sec:experiments}, we conduct experiments to validate the superiority of our method in comparison to the state-of-the-art attack and defense models. Finally, Section \ref{sec:conclusion} concludes this paper.

\section{Related Work}\label{sec:related_work}
\subsection{Deep Hashing-based Retrieval}
Hashing methods aim to learn a hash function to convert semantically similar samples into similar hash codes in Hamming space, which are widely used to accelerate ANN retrieval \cite{andoni2006near}.  Particularly, deep hashing \cite{luo2023survey} utilizes deep neural networks as hash functions for feature extraction and binary code generation in an end-to-end manner, attaining superior retrieval performance. Existing deep hashing can be broadly categorized into two primary streams: unsupervised and supervised deep hashing. Unsupervised deep hashing methods \cite{salakhutdinov2009semantic,ghasedi2018unsupervised} usually involve learning hash functions by mining inherent structural or distribution-relevance similarities in the samples without using any semantic label. By contrast, supervised deep hashing methods use semantic labels or relevant information as supervisory signals to overcome the semantic gap dilemma \cite{smeulders2000content}, which can yield more precise performance than unsupervised ones. As the first deep hashing algorithm, CNNH \cite{xia2014supervised} consists of two independent stages, \textit{i.e.}, designing approximate hash codes of training data and learning feature representation through DNN. 
Recent hashing methods \cite{li2016feature,cao2017hashnet,yuan2020central,wang2020deep,fan2020deep,hoe2021one,doan2022one,wu2023deep} focused on the design of end-to-end strategies and loss functions to improve the efficacy of hashing learning. For example, DPSH \cite{li2016feature} integrated image representation and hash coding in a unified framework and adopted a pairwise loss to preserve the semantic similarity between data objects. HashNet \cite{cao2017hashnet} proposed a continuous scale strategy to tackle the optimization problem in discrete Hamming space, and alleviated the data imbalance by a weighted pairwise loss. CSQ\cite{yuan2020central} and DPN\cite{fan2020deep} are devoted to finding class-wise hash representatives (centers) that can provide global similarity for hashing learning. Besides, HSWD \cite{doan2022one} presented a Sliced-Wasserstein-based distributional distance to achieve low quantization error and coding balance.

\subsection{Adversarial Attack}
As the vulnerability of deep learning models has been noticed by Szegedy \textit{et al.} \cite{szegedy2013intriguing}, numerous studies have explored the model’s robustness and proposed a series of attack methods. Depending on the target model information available to the attackers, adversarial attacks can be divided into white-box attacks \cite{szegedy2013intriguing,goodfellow2014explaining,moosavi2016deepfool,madry2017towards,kurakin2018adversarial,lin2019nesterov,croce2020reliable,xu2022bounded} and black-box attacks\cite{zhong2020towards, xie2019improving,wang2021enhancing,yuan2022adaptive}. For white-box attacks, the whole network architecture and parameters are exposed to attackers, so that they can optimize the adversarial perturbations according to the gradients of the loss \textit{w.r.t.} inputs. For instance, FGSM \cite{goodfellow2014explaining}
is a classic white-box attack method, which crafts adversarial samples by maximizing the loss along the gradient direction with a large step. After that, FGSM was further extended to multi-step variants, such as BIM \cite{kurakin2018adversarial} and PGD \cite{madry2017towards}.
In addition, a simple yet accurate method named Deepfool \cite{moosavi2016deepfool} was developed to generate minimal perturbations sufficient to change classification labels. Recently, some methods \cite{croce2020reliable,xu2022bounded} with stronger attack capabilities have been proposed, posing a greater threat to the robustness of deep learning models.
For black-box attacks, attackers can only access the inputs and outputs of target models, so it is difficult to acquire the gradient directly. One popular solution is to exploit the transferability of adversarial example to conduct the attack \cite{xie2019improving,wang2021enhancing,yuan2022adaptive}.

Apart from classification, researchers have also explored adversarial attacks on deep hashing-based retrieval \cite{yang2018adversarial,bai2020targeted,wang2021prototype,wang2021targeted,lu2021smart,zhang2021targeted}. HAG \cite{yang2018adversarial} is the first attack method in this field, which can confuse hashing models to retrieve results irrelevant to the input sample, \textit{i.e.}, \textbf{non-targeted attack}. Subsequently, Lu \textit{et al.} \cite{lu2021smart} proposed SDHA to obtain more effective adversarial queries against retrieval tasks by considering staying away from all relevant samples of the query. For the \textbf{targeted attack} (\textit{i.e.}, the retrieved images of the adversarial example are semantically relevant to the target label specified by the attacker), P2P and DHTA heuristically \cite{bai2020targeted} select an anchor code as the representative of the target label and then move the adversarial query close to it. Recently, ProS-GAN \cite{wang2021prototype} and THA \cite{wang2021targeted} design an auxiliary network to produce prototype code as a guide for targeted attacks and defense. Unlike the auxiliary network without theoretical guarantees, we propose a discriminative learning measurement with a provable mathematical formula to obtain the representative mainstay codes.

\subsection{Adversarial Training}
To resist the adversarial examples, many defense methods \cite{xie2017mitigating,jia2019comdefend,madry2017towards,zhang2019theoretically,wu2020adversarial,cui2021learnable,jia2022adversarial} have been proposed. Among them, adversarial training is currently the most effective way to strengthen the robustness of neural networks, which augments the training data with adversarial examples. Madry \textit{et al.} \cite{madry2017towards} reformulated standard adversarial training as a minimax optimization problem. After that, Zhang \textit{et al.} \cite{zhang2019theoretically} proposed a defense method named TRADES, which provides a trade-off between adversarial robustness and benign accuracy on clean data. Cui \textit{et al.} \cite{cui2021learnable} exploited the natural classifier boundary as a guide to improve model robustness without losing much natural accuracy. Jia \textit{et al.} \cite{jia2022adversarial} introduced a learnable attack strategy for adversarial training. In detail, they designed a strategy network to automatically produce sample-dependent attack strategies at different training stages.

In deep hashing, Wang \textit{et al.} \cite{wang2021targeted} proposed an adversarial training algorithm based on the targeted attack (dubbed ATRDH here) by reconstructing the semantic correlations between adversarial samples and clean samples. It is clear that ATRDH is not a standard adversarial training mechanism, and it simply treats the similarity errors induced by the adversarial samples as a regularization term. By contrast, our SAAT is a standardized adversarial training that minimizes the distance between the hash codes of adversarial examples and the mainstay codes under a well-designed \textit{minimax} framework.

\section{Semantic-Aware Adversarial Training}\label{sec:method}
This section will present the proposed Semantic-Aware Adversarial Training (SAAT) framework, which is a standardized minimax optimization formulation for deep hashing aiming to achieve robust deep hashing models. We first present the definitions of adversarial attack and adversarial training on deep hashing-based retrieval and then explicitly elaborate on the overall idea and submodules, followed by a discussion.

\begin{table}[ht]
\small
\centering
\caption{Summary of main notation.}
\label{tab:notation}
\resizebox{0.48\textwidth}{!}{
\begin{tabular}{c|c}
\hline
Notation & Description \\
\hline
$F(\cdot)$ & hashing model \\
$f_{\theta}(\cdot)$ & DNN with parameter $\theta$ \\
$O$ & training set, $O=\{(\boldsymbol{x}_i,\boldsymbol{y}_i)\}_{i=1}^N$ \\
$\boldsymbol{x}, \boldsymbol{y}$ & input image and its corresponding label \\
$\boldsymbol{x^{\prime}}$ & the adversarial example of $\boldsymbol{x}$ \\
$\boldsymbol{b}$ & the hash code of $\boldsymbol{x}$ \\
$S$ & similarity matrix \\
$\boldsymbol{x}_{\bar{i}}^{(\rm{p})}$, $\boldsymbol{x}_{\bar{j}}^{(\rm{n})}$ & the $i$-th positive and $j$-th negative of $\boldsymbol{x}$ \\
$\boldsymbol{b}_{\bar{i}}^{(\rm{p})}$, $\boldsymbol{b}_{\bar{j}}^{(\rm{n})}$ & the hash codes of $\boldsymbol{x}_{\bar{i}}^{(\rm{p})}$ and $\boldsymbol{x}_{\bar{j}}^{(\rm{n})}$ \\
$w_{\bar{i}}$, $w_{\bar{j}}$ & the weighting coefficients for $\boldsymbol{x}_{\bar{i}}^{(\rm{p})}$ and $\boldsymbol{x}_{\bar{j}}^{(\rm{n})}$ \\
$\boldsymbol{b}_{m}$ & the mainstay code of $\boldsymbol{x}$\\
$\boldsymbol{b}_{t}$ & the mainstay code of target label $\boldsymbol{y}_{t}$\\
$D_{H}$ & Hamming distance measure function \\
$N$ & the number of training samples \\
$C$ & the number of classes \\
$K$ & the length of hash code \\
$N_{\rm{p}}$, $N_{\rm{n}}$ & the numbers of positives and negatives of $\boldsymbol{x}$ \\
$\mathcal{L}_{adv}$ & objective function for adversarial attack \\
$\mathcal{L}_{at}$ & objective function for adversarial training \\
\hline
\end{tabular}
}
\end{table}

\subsection{Preliminaries}
\noindent \textbf{Deep Hashing-based Retrieval.} 
We consider learning a hashing model $F$ from a training set $O=\{(\boldsymbol{x}_i,\boldsymbol{y}_i)\}_{i=1}^N$ that contains $N$ samples labeled with $C$ classes, where $\boldsymbol{x}_i$ indicates $i$-th image, and $\boldsymbol{y}_i=[y_{i1},y_{i2},...,y_{iC}]\in \{0,1\}^C$ denotes a label vector of $\boldsymbol{x}_i$. $y_{ij}=1$ means that $\boldsymbol{x}_i$ belongs to the $j$-th class. It is worth noting that $\boldsymbol{x}_i$ is allowed to belong to more than one class, \textit{i.e.}, multi-label data. The objective of $F$ is to get a set of $K$-bit binary codes $B = \{\boldsymbol{b}_i\}_{i=1}^N \in \{-1, 1\}^{N \times K}$ for the training set, which desires to preserve semantic similarities among samples in Hamming space for efficient ANN search. Generally, we utilize similarity matrix $S$ to express semantic similarities between each pair of samples. For any two instances $\boldsymbol{x}_i$ and $\boldsymbol{x}_j$, $S_{ij} > 0$ describes they share at least one class, otherwise $S_{ij} = 0$.

The generation process of hash code $\boldsymbol{b}_i$ of $\boldsymbol{x}_i$ can be expressed as follows:
\begin{equation}
    \begin{aligned}
        &\boldsymbol{b}_i = {F}(\boldsymbol{x}_i) = \operatorname{sign}(f_\theta(\boldsymbol{x}_i)), \ \text{s.t.~}  \boldsymbol{b}_i \in\{-1,1\}^K,
    \end{aligned}
\end{equation}
where $K$ represents the hash code length, and $f(\cdot)$ with parameter $\theta$ is a DNN to approximate hash function ${F}(\cdot)$. The final binary code $\boldsymbol{b}_i$ is obtained by applying the $\operatorname{sign}(\cdot)$ on the output of $f_\theta(\boldsymbol{x}_i)$. Typically, $f(\cdot)$ is implemented by a convolutional neural network (CNN) and adopts $\operatorname{tanh}(\cdot)$ function to approximate the $\operatorname{sign}(\cdot)$ function during the training process to relieve the vanishing gradient problem.

\noindent \textbf{Definition 1: \textit{Adversarial Attack on Deep Hashing-based Retrieval}.} In deep hashing-based retrieval, given a benign query $\boldsymbol{x}$ with label $\boldsymbol{y}$, the goal of non-targeted attack is to craft an adversarial example $\boldsymbol{x}^{\prime}$, which could confuse the deep hashing model $F$ to retrieve irrelevant samples to query $\boldsymbol{x}$. In contrast, a targeted attack aims to mislead the deep hashing model into returning samples related to a given target label $\boldsymbol{y}_{t}$. Moreover, the adversarial perturbation $\boldsymbol{x}^{\prime} - \boldsymbol{x}$ should be as small as possible to be imperceptible to the human eye.

\noindent \textbf{Definition 2: \textit{Adversarial Training on Deep Hashing-based Retrieval}.}
Similar to classification, adversarial training on deep hashing utilizes both the benign samples $\{(\boldsymbol{x}_i,\boldsymbol{y}_i)\}_{i=1}^N$ and corresponding adversarial versions $\{(\boldsymbol{x}_i^{\prime},\boldsymbol{y}_i)\}_{i=1}^N$ to re-optimize the parameter $\theta$ of deep hashing model, and thereby the model could retrieve semantically relevant contents to the original label $\boldsymbol{y}_i$, whether the input is a clean sample $\boldsymbol{x}_i$ or an adversarial sample $\boldsymbol{x}_i^{\prime}$.

\noindent \textbf{Summary of Notation.} For the sake of clarity, we summarize the formal notation and key concepts in Table \ref{tab:notation}.

\subsection{An Overall Illustration}
During the procedure of adversarial learning, the generation of adversarial examples plays a key role. Unlike image classification with labels as supervision to generate practical adversarial examples, deep hashing suffers from a lack of discriminative representatives to guide adversarial attack and defense. To address this issue, we conceive a semantic-aware code (\textit{i.e.}, mainstay code) for each class as an optimal representative of adversarial learning. With the mainstay code, we present an efficient adversarial attack method and formulate a well-designed adversarial training based on a minimax scheme for deep hashing.
The overall framework of our proposed SAAT contains two components: the generation of mainstay code and the minimax-based adversarial training, as shown in Fig. \ref{fig:framework}. Particularly, we construct a mainstay code for each class by a discriminative mainstay features learning pattern that ensures the mainstay code is as close as possible to the positives and stays away from the negatives. After yielding the mainstay codes, we serve a minimax-based adversarial training. Under formalized adversarial training, the inner maximization aims to generate adversarial examples led by the mainstay codes, while the outer minimization attempts to learn robust models by minimizing the expected loss over the adversarial perturbations.

\subsection{Generation of Mainstay Code}
We desire to build a semantic-aware representative (dubbed \textit{mainstay code}) for any class in Hamming space to accomplish adequate adversarial attacks and learn robust deep hashing models. Considering the specialties of deep hashing, we suggest that the mainstay code is required to satisfy the following properties. 1) First, the mainstay code is a binary code in Hamming Space with length $K$ to retain the computation efficiency of hashing, where $K$ is the hash code length of given deep hashing models. 2) Second, the mainstay code can be generated adaptively for different datasets and multiple deep hashing methods. 3) More importantly, the mainstay code is expected to be globally semantically discriminable, \textit{i.e.}, as close as possible to all positives of the given class and as far as possible from all negatives of the class in Hamming space, as illustrated in Fig. \ref{fig:framework}. Notably, for any sample $\boldsymbol{x}_i$ with class $\boldsymbol{y}_i$, its corresponding positives are those share at least one class with itself, \textit{i.e.}, $\{ (\boldsymbol{x}_j, \boldsymbol{y}_j) \in O : S_{ij} > 0\}$, and its corresponding negatives are $\{ (\boldsymbol{x}_j, \boldsymbol{y}_j) \in O : S_{ij} = 0\}$. 

Subsequently, we will present the well-conceived discriminative mainstay features learning (DMFL) scheme to solve the mainstay code. For conciseness, we will discard the subscript $i$ and just use $\boldsymbol{x}$ labeled with $\boldsymbol{y}$ to denote the input sample. In particular, we treat $\bar{i}$ and $\bar{j}$ as the indices of positives and negatives of $\boldsymbol{x}$, respectively. That is, $\boldsymbol{x}_{\bar{i}}^{(\rm{p})}$ is the $i$-th positive sample of $\boldsymbol{x}$, and $\boldsymbol{x}_{\bar{j}}^{(\rm{n})}$ is the $j$-th negative sample of $\boldsymbol{x}$.
According to the third property, the mainstay code $\boldsymbol{b}_m$ of a given sample $(\boldsymbol{x}, \boldsymbol{y})$ can be formulated as follows: 
\begin{equation}
	\begin{aligned}
	    \min_{\boldsymbol{b}_{m}} \sum_{\bar{i}}^{N_{\rm{p}}} w_{\bar{i}} {D}_{\rm{H}}(\boldsymbol{b}_{m}, \boldsymbol{b}_{\bar{i}}^{(\rm{p})}) - \sum_{\bar{j}}^{N_{\rm{n}}} w_{\bar{j}} {D}_{\rm{H}}(\boldsymbol{b}_{m}, \boldsymbol{b}_{\bar{j}}^{(\rm{n})})
	\end{aligned},
\label{eq:optimal_solution}
\end{equation}
where $D_{\rm{H}}$ is the Hamming distance measure. $\boldsymbol{b}_{\bar{i}}^{(\rm{p})}$ and $\boldsymbol{b}_{\bar{j}}^{(\rm{n})}$ are hash codes of $\boldsymbol{x}_{\bar{i}}^{(\rm{p})}$ and $\boldsymbol{x}_{\bar{j}}^{(\rm{n})}$, respectively. $N_{\rm{p}}$ and $N_{\rm{n}}$ are the numbers of positive and negative samples, respectively. $w_{\bar{i}}$ and $w_{\bar{j}}$ are the weighting coefficients for different samples.

However, generating and optimizing the mainstay code are challenging to achieve the optimal solution, due to the discrete nature of the mainstay code and high computational cost. Fortunately, based on the binary characteristic of the hash codes, we can precisely figure out the mainstay code by the following theorem.

~\\
\noindent \textbf{Theorem 1.} \textit{
Suppose $\boldsymbol{b} \in \{-1, +1\}^{K}$ is a binary code in Hamming space, and $\psi(\boldsymbol{b})$ is an objective function as follows:
\begin{equation}
    \begin{aligned}
    \psi(\boldsymbol{b})=\sum_{\bar{i}}^{N_{\rm{p}}} w_{\bar{i}} {D}_{\rm{H}}(\boldsymbol{b}, \boldsymbol{b}_{\bar{i}}^{(\rm{p})}) - \sum_{\bar{j}}^{N_{\rm{n}}} w_{\bar{j}} {D}_{\rm{H}}(\boldsymbol{b}, \boldsymbol{b}_{\bar{j}}^{(\rm{n})})
    \end{aligned}.
\end{equation}
If $\boldsymbol{b}_{m}$ is the optimal solution to $\min\psi(\boldsymbol{b})$, then $\boldsymbol{b}_{m}$ can be directly written as
\begin{equation}
    \begin{aligned}
    \boldsymbol{b}_{m} &= \arg\min_{\boldsymbol{b}\in\{-1,+1\}^K} \sum_{\bar{i}}^{N_{\rm{p}}} w_{\bar{i}} {D}_{\rm{H}}(\boldsymbol{b}, \boldsymbol{b}_{\bar{i}}^{(\rm{p})}) - \sum_{\bar{j}}^{N_{\rm{n}}} w_{\bar{j}} {D}_{\rm{H}}(\boldsymbol{b}, \boldsymbol{b}_{\bar{j}}^{(\rm{n})}) \\
    &=\operatorname{sign}\left(\sum_{\bar{i}}^{N_{\rm{p}}}w_{\bar{i}}\boldsymbol{b}_{\bar{i}}^{(\rm{p})} - \sum_{\bar{j}}^{N_{\rm{n}}}w_{\bar{j}}\boldsymbol{b}_{\bar{j}}^{(\rm{n})} \right).
    \end{aligned}
\end{equation}
}

\noindent \textit{Proof.} As the mainstay code $\boldsymbol{b}_{m}$ is the optimal solution of the minimizing objective, the above theorem is equivalent to prove the following inequality:
\begin{equation}
    \begin{aligned}
    \psi(\boldsymbol{b})\geq \psi(\boldsymbol{b}_{m}),
    \quad \forall~\boldsymbol{b}\in\{-1,+1\}^K
    \end{aligned}.
\end{equation}
According to $D_{\rm{H}}(\boldsymbol{b}_1, \boldsymbol{b}_2)=\frac{1}{2}(K-\boldsymbol{b}_1^{\top}\boldsymbol{b}_2)$ and $\boldsymbol{b}=\{b_1,b_2,...,b_K\}$,  then we have
\begin{equation}
    \begin{aligned}
    &\psi(\boldsymbol{b})
    =\sum_{\bar{i}}^{N_{\rm{p}}} w_{\bar{i}} \frac{1}{2}(K-\boldsymbol{b}^\top \boldsymbol{b}_{\bar{i}}^{(\rm{p})}) - \sum_{\bar{j}}^{N_{\rm{n}}} w_{\bar{j}} \frac{1}{2}(K-\boldsymbol{b}^\top \boldsymbol{b}_{\bar{j}}^{(\rm{n})})\\
    =&-\frac{1}{2}\sum_{\bar{i}}^{N_{\rm{p}}} w_{\bar{i}}\boldsymbol{b}^\top \boldsymbol{b}_{\bar{i}}^{(\rm{p})}+\frac{1}{2}\sum_{\bar{j}}^{N_{\rm{n}}} w_{\bar{j}}\boldsymbol{b}^\top \boldsymbol{b}_{\bar{j}}^{(\rm{n})} + \xi \\
    =&-\frac{1}{2}\sum_{\bar{i}}^{N_{\rm{p}}}w_{\bar{i}}\sum_{k=1}^K{b}_k{b}_{\bar{i}k}^{(\rm{p})}+\frac{1}{2}\sum_{\bar{j}}^{N_{\rm{n}}} w_{\bar{j}}\sum_{k=1}^K{b}_k{b}_{\bar{j}k}^{(\rm{p})}+\xi \\
    =&-\frac{1}{2}\sum_{k=1}^K b_k(\sum_{\bar{i}}^{N_{\rm{p}}}w_{\bar{i}}b_{\bar{i}k}^{(\rm{p})}-\sum_{\bar{j}}^{N_{\rm{n}}}w_{\bar{j}}b_{\bar{j}k}^{(\rm{n})})+\xi, \\
    \end{aligned}
    \label{eq:psi_b}
\end{equation}
where $\xi$ is a constant. Due to the nature of absolute value, we have
\begin{equation}
    \begin{aligned}
    &\psi(\boldsymbol{b})
    =-\frac{1}{2}\sum_{k=1}^K b_k(\sum_{\bar{i}}^{N_{\rm{p}}}w_{\bar{i}}b_{\bar{i}k}^{(\rm{p})}-\sum_{\bar{j}}^{N_{\rm{n}}}w_{\bar{j}}b_{\bar{j}k}^{(\rm{n})})+\xi \\
    \geq&-\frac{1}{2}\sum_{k=1}^K \left|b_k(\sum_{\bar{i}}^{N_{\rm{p}}}w_{\bar{i}}b_{\bar{i}k}^{(\rm{p})}-\sum_{\bar{j}}^{N_{\rm{n}}}w_{\bar{j}}b_{\bar{j}k}^{(\rm{n})})\right|+\xi \\
    =&-\frac{1}{2}\sum_{k=1}^K \left|(\sum_{\bar{i}}^{N_{\rm{p}}}w_{\bar{i}}b_{\bar{i}k}^{(\rm{p})}-\sum_{\bar{j}}^{N_{\rm{n}}}w_{\bar{j}}b_{\bar{j}k}^{(\rm{n})})\right|+\xi \\
    =&-\frac{1}{2}\sum_{k=1}^K \operatorname{sign}(\sum_{\bar{i}}^{N_{\rm{p}}}w_{\bar{i}}b_{\bar{i}k}^{(\rm{p})}-\sum_{\bar{j}}^{N_{\rm{n}}}w_{\bar{j}}b_{\bar{j}k}^{(\rm{n})}) (\sum_{\bar{i}}^{N_{\rm{p}}}w_{\bar{i}}b_{\bar{i}k}^{(\rm{p})}\\ &-\sum_{\bar{j}}^{N_{\rm{n}}}w_{\bar{j}}b_{\bar{j}k}^{(\rm{n})})+\xi
    \end{aligned}
\end{equation}
Similar to Eq. (\ref{eq:psi_b}), we represent $\psi(\boldsymbol{b}_{m})$ as
\begin{equation}
    \begin{aligned}
    \psi(\boldsymbol{b}_{m})=-\frac{1}{2}\sum_{k=1}^K b_{mk}(\sum_{\bar{i}}^{N_{\rm{p}}}w_{\bar{i}}b_{\bar{i}k}^{(\rm{p})}-\sum_{\bar{j}}^{N_{\rm{n}}}w_{\bar{j}}b_{\bar{j}k}^{(\rm{n})})+\xi.
    \end{aligned}
\end{equation}
Hence, we have 
\begin{align}
\psi(\boldsymbol{b})
\geq & -\frac{1}{2}\sum_{k=1}^K \operatorname{sign}(\sum_{\bar{i}}^{N_{\rm{p}}}w_{\bar{i}}b_{\bar{i}k}^{(\rm{p})}-\sum_{\bar{j}}^{N_{\rm{n}}}w_{\bar{j}}b_{\bar{j}k}^{(\rm{n})}) (\sum_{\bar{i}}^{N_{\rm{p}}}w_{\bar{i}}b_{\bar{i}k}^{(\rm{p})}\nonumber\\ &-\sum_{\bar{j}}^{N_{\rm{n}}}w_{\bar{j}}b_{\bar{j}k}^{(\rm{n})})+\xi\nonumber\\
=&-\frac{1}{2}\sum_{k=1}^K b_{mk}(\sum_{\bar{i}}^{N_{\rm{p}}}w_{\bar{i}}b_{\bar{i}k}^{(\rm{p})}-\sum_{\bar{j}}^{N_{\rm{n}}}w_{\bar{j}}b_{\bar{j}k}^{(\rm{n})})+\xi \\
=&\psi(\boldsymbol{b}_{m}).\nonumber
\end{align}

That is, $\psi(\boldsymbol{b})\geq\psi(\boldsymbol{b}_m)$ and the theorem is proved. Therefore, the mainstay code $\boldsymbol{b}_m$ of a given sample $(\boldsymbol{x}, \boldsymbol{y})$ in problem (\ref{eq:optimal_solution}) can be solved by 
\begin{equation}
    \begin{aligned}
    \boldsymbol{b}_m =\operatorname{sign}\left(\sum_{\bar{i}}^{N_{\rm{p}}}w_{\bar{i}}\boldsymbol{b}_{\bar{i}}^{(\rm{p})} - \sum_{\bar{j}}^{N_{\rm{n}}}w_{\bar{j}}\boldsymbol{b}_{\bar{j}}^{(\rm{n})} \right).
    \end{aligned}
\label{eq:mainstay_code}
\end{equation}
In particular, we have to claim that our discriminative mainstay code is a globally optimal semantic representative. Compared to sample-level \cite{yang2018adversarial, bai2020targeted} and category-level \cite{bai2020targeted, lu2021smart} approaches, our mainstay code is optimized in global semantic space by preserving similarity with all positives and irrelevancy with all negatives. Here, “globally” means that our method considers all positive and negative samples related to the query in the global semantic space, not that our mainstay code can be the optimal semantic representative in all cases.

In addition, we define the $w_{\bar{i}}$ and $w_{\bar{j}}$ as follows:
\begin{equation}
    w_{\bar{i}} = \frac{1}{N_{\rm{p}}} \cdot s_{\bar{i}}, \quad w_{\bar{j}} = \frac{1}{N_{\rm{n}}} \cdot (1 - s_{\bar{j}}),
\end{equation}
where $s_{\bar{i}/\bar{j}} = \frac{\left\langle \boldsymbol{y}, \boldsymbol{y}_{\bar{i}/\bar{j}}  \right\rangle }{\left\|\boldsymbol{y}\right\| \left\|\boldsymbol{y}_{\bar{i}/\bar{j}}\right\|}$ denotes the similarity between the given instance $(\boldsymbol{x}, \boldsymbol{y})$ and the $i$-th positive ($j$-th negative), which means the more classes they share, the more similar they are. $\bar{i}/\bar{j}$ indicates $\bar{i}$ `or' $\bar{j}$. $\frac{1}{N_{\rm{p/n}}}$ can balance the number difference between positive and negative samples.

\subsection{Semantic-Aware Adversarial Attack} \label{sec:adversarial_attack}
To verify the effectiveness of our mainstay code and reveal the robustness of deep hashing models, we provide a semantic-aware adversarial attack strategy based on proposed mainstay code. We first illustrate the implementation of our approach for non-targeted attack, and then we apply the idea to targeted attack. 

\noindent \textbf{Non-targeted Attack.} For non-targeted attack, we prefer maximizing the hash code distance between the adversarial example and its semantically relevant samples, and simultaneously minimizing the distance from irrelevant samples, rather than the only benign sample. Since the mainstay code is found, the objective of adversarial attack can be translated into maximizing the Hamming distance between the hash code of adversarial example and the mainstay code.
Thus, for a given clean image $\boldsymbol{x}$, its corresponding adversarial example $\boldsymbol{x}^\prime$ is developed by following objective under the $L_{p}$ constraint:
\begin{equation}
    \begin{aligned}
        \boldsymbol{x}^{\prime} = \arg\max_{\boldsymbol{x}^\prime} D_{\rm{H}} (F(\boldsymbol{x}^\prime), \boldsymbol{b}_{m}), \quad 
        \text{s.t.~} \|\boldsymbol{x} & - \boldsymbol{x}^\prime\|_{p} \leq \epsilon,
    \end{aligned}
    \label{eq:non-targeted_attack}
\end{equation}
where $\|\cdot\|_{p}$ ($p={1,2,\infty}$) is $L_p$ norm that keeps the pixel difference between the adversarial example and the benign sample no more than $\epsilon$ for the imperceptible property of adversarial perturbations.

Due to $D_{\rm{H}}(\boldsymbol{b}_1, \boldsymbol{b}_2)=\frac{1}{2}(K-\boldsymbol{b}_1^{\top}\boldsymbol{b}_2)$, the Eq. (\ref{eq:non-targeted_attack}) is equivalent to:
\begin{equation}
    \begin{aligned}
        \boldsymbol{x}^{\prime} &= \arg\max_{\boldsymbol{x}^\prime} -\frac{1}{K}\boldsymbol{b}_{m}^{\top}\operatorname{tanh}(\alpha f_{\theta}(\boldsymbol{x}^\prime)), \\
        & \text{s.t.~} \|\boldsymbol{x} - \boldsymbol{x}^\prime\|_{p} \leq \epsilon,
    \end{aligned}
    \label{eq:att}
\end{equation}
where $\operatorname{tanh}(\cdot)$ is the tanh activation function, and $\alpha \in [0, 1]$ is the hyper-parameter that controls $\operatorname{tanh}(\alpha f_{\theta}(\boldsymbol{x}^\prime)$ to approximate $F(\boldsymbol{x}^\prime)$. Following \cite{yang2018adversarial}, we initialize $\alpha$ with a small value to obtain a larger gradient of the objective function in Eq. (\ref{eq:att}) concerning $\boldsymbol{x}^\prime$, then gradually enlarge it to approximate $\operatorname{sign}(\cdot)$ until it equals $1$. 

\noindent \textbf{Targeted Attack.} The only difference between non-targeted attacks and targeted attacks is the objective function. For a given benign sample $\boldsymbol{x}$ and a target label $\boldsymbol{y}_{t}$, we first acquire the mainstay code $\boldsymbol{b}_t$ of $\boldsymbol{y}_t$ by Eq. (\ref{eq:mainstay_code}), and then the objective of targeted attack can be defined as:
\begin{equation}
    \begin{aligned}
        \boldsymbol{x}^{\prime} &= \arg\min_{\boldsymbol{x}^\prime} D_{\rm{H}} (F(\boldsymbol{x}^\prime), \boldsymbol{b}_t) \\
        &=\arg\max_{\boldsymbol{x}^\prime} \frac{1}{K}\boldsymbol{b}_{t}^{\top}\operatorname{tanh}(\alpha f_{\theta}(\boldsymbol{x}^\prime)), \\
        & \text{s.t.~} \|\boldsymbol{x} - \boldsymbol{x}^\prime\|_{p} \leq \epsilon.
    \end{aligned}
    \label{eq:targeted_attack}
\end{equation}
As the distance between the hash code of adversarial example $\boldsymbol{x}^{\prime}$ and the mainstay code of the target label decreases, the adversarial example gradually approaches the target label in semantic terms while guaranteeing visual imperceptibility. Thus, we could retrieve semantically relevant contents to the target label by feeding $\boldsymbol{x}^{\prime}$ into a deep hashing-based retrieval system.

\noindent \textbf{Generation of Adversarial Examples.} Given a clean image, this paper adopts PGD attack \cite{madry2017towards}, one of the most popular attack methods (other attack strategies are also feasible), to optimize $\boldsymbol{x}^\prime$ with $T$ ($T=100$ by default) iterations, \textit{i.e.},
\begin{equation}
    \begin{aligned}
        \boldsymbol{x}_T^\prime = {\mathcal{S}}_{\epsilon}(\boldsymbol{x}_{T-1}^\prime + \eta \cdot \operatorname{sign}(\nabla_{\boldsymbol{x}_{T-1}^\prime}\mathcal{L}_{adv})),
		\quad \boldsymbol{x}^{\prime}_0 = \boldsymbol{x},
    \end{aligned}
\end{equation}
where $\eta$ is the step size, $\mathcal{S}_{\epsilon}$ projects $\boldsymbol{x}^\prime$ into the $\epsilon$-ball \cite{madry2017towards} of $x$, and $\mathcal{L}_{adv}$ is the objective function for adversarial attack. In the case of non-targeted attack, $\mathcal{L}_{adv} = -\frac{1}{K}\boldsymbol{b}_{m}^{\top}\operatorname{tanh}(\alpha f_{\theta}(\boldsymbol{x}^\prime))$, while $\mathcal{L}_{adv} = \frac{1}{K}\boldsymbol{b}_{t}^{\top}\operatorname{tanh}(\alpha f_{\theta}(\boldsymbol{x}^\prime))$ for targeted attack. 

Unlike HAG, SDHA and DHTA which require more than $1000$ iterations to optimize adversarial examples, our method is more efficient. It merely takes $100$ iterations because of the well-designed globally optimal semantic representatives. Notably, we should highlight that since the generation process of adversarial examples is model-agnostic, our learning algorithm could be simply embedded into any deep pairwise similarity-preserving hashing network, such as DPSH\cite{li2016feature}, HashNet \cite{cao2017hashnet}, CSQ\cite{yuan2020central}, etc.

\subsection{Semantic-Aware Adversarial Training}
The ultimate pursuit of adversarial learning is to enhance the robustness of deep neural networks. After the powerful adversarial attack materializes, we further aspire that the produced adversarial examples can be used as augmentation data to optimize the deep hashing model for defense, \textit{i.e.}, adversarial training. According to \cite{madry2017towards}, the standard adversarial training on classification can be written as a minimax formulation, \textit{i.e.},
\begin{equation}
    \begin{aligned}
        \min_{\theta} \mathbb{E}_{(\boldsymbol{x}, \boldsymbol{y}) \sim \mathcal{D}} [\max_{\boldsymbol{x^{\prime}}} \mathcal{J}(g_{\theta}(\boldsymbol{x}^{\prime}), y, \theta)],
    \end{aligned}
    \label{eq:minimax}
\end{equation}
where $\mathcal{D}$ represents an underlying data distribution, $g$ is a classifier with parameter $\theta$, and $\mathcal{J}$ denotes a classification loss (\textit{e.g.}, cross-entropy loss). The inner maximization problem in (\ref{eq:minimax}) can be regarded as an adversarial attack that finds an adversarial example, and the outer optimizes the parameter of network to resist the influence caused by adversarial perturbations. Generally, such a framework is unavailable in deep hashing due to the absence of explicit semantic representative (\textit{e.g.}, label). Nevertheless, the proposed mainstay code remedies this deficiency, thereby enabling us to develop semantic-aware adversarial training (SAAT) for deep hashing, \textit{i.e.},
\begin{equation}
    \begin{aligned}
        \min_{\theta} \mathbb{E}_{(\boldsymbol{x}, \boldsymbol{y}) \sim \mathcal{D}} [\max_{\boldsymbol{x}^\prime}\mathcal{L}_{at}(\boldsymbol{x}, \boldsymbol{x}^{\prime}, \boldsymbol{b}_m, \theta)]
    \end{aligned}.
    \label{eq:minimax_saat}
\end{equation}

The developed SAAT has two significant differences compared to the regular adversarial training in classification. First, the most essential difference is that we leverage the representative mainstay code $\boldsymbol{b}_m$ to guide the adversarial training instead of label vector $y$. In addition, the objective function in internal optimization is different. The optimization objective $\mathcal{L}_{at}$ of SAAT is more complex, which contains three loss items.
In the first item, we exploit the semantic-aware adversarial attack (non-targeted attack) described in Section \ref{sec:adversarial_attack} as the attack strategy to seek the optimal adversarial example. Thus, the first loss item of $\mathcal{L}_{at}$ is formulated as follows:
\begin{equation}
    \begin{aligned}
        \mathcal{L}_{adv}(\boldsymbol{x}^{\prime}, \boldsymbol{b}_{m}; \theta)=-\frac{1}{K}\boldsymbol{b}_{m}^{\top}\operatorname{tanh}(f_{\theta}(\boldsymbol{x}^\prime))
    \end{aligned}.
\end{equation}
To make the back-propagation algorithm feasible during training, we substitute the sign function with the tanh function to obtain approximate continuous hash codes, which causes quantization errors. Thus, we introduce a quantization loss to minimize the discrepancy between the approximate hash codes and the binary codes of adversarial examples, \textit{i.e.},
\begin{equation}
    \begin{aligned}
        \mathcal{L}_{qua} (\boldsymbol{x}^{\prime}; \theta) =\|\operatorname{tanh}(f_{\theta}(\boldsymbol{x}^\prime))-\operatorname{sign}(f_{\theta}(\boldsymbol{x}^\prime))\|_2^2
    \end{aligned},
\end{equation}
where $\|\cdot\|_2$ is the $L_2$ norm. It is worth noting that an excellent adversarial training strategy should not only improve the robustness of models against adversarial examples but also maintain the performance on benign samples. Hence, we also employ the objective function $\mathcal{L}_{ori} (\boldsymbol{x}, \theta)$ of the original hashing method (\textit{e.g.}, DPH, DPSH or HashNet).
In summary, the whole objective function of SAAT is described as:
\begin{equation}
    \begin{aligned}
        \mathcal{L}_{at}=\lambda\mathcal{L}_{adv}+\mu\mathcal{L}_{qua}+\mathcal{L}_{ori},
    \end{aligned}
    \label{eq:obj_at}
\end{equation}
where $\lambda$ and $\mu$ are the trade-off hyper-parameters of the first two terms.

\renewcommand{\algorithmicrequire}{\textbf{Input:}}
\renewcommand{\algorithmicensure}{\textbf{Output:}}
\begin{algorithm}[t]
	\caption{Optimization of Semantic-Aware Adversarial Training in Problem (\ref{eq:minimax_saat})}
	\label{alg:saat}
	\begin{algorithmic}[1]
		\REQUIRE
		Image dataset $O=\{(\boldsymbol{x}_i, \boldsymbol{y}_i)\}_{i=1}^N$, pre-trained hashing model $F(\cdot)=\operatorname{sign}(f_\theta(\cdot))$, training epochs $E$, batch size $n$, learning rate $\zeta$, step size $\alpha$, perturbation budget $\epsilon$, attack iterations $T$, weighting factors $\lambda$ and $\mu$.
		\FOR{$iter=1...E$}
		\FOR{image batch $\{(x_i, y_i)\}_{i=1}^n$}
		\STATE For each $x_i$, calculate its mainstay code $\boldsymbol{b}_{mi}$ with Eq. (\ref{eq:mainstay_code});
		\STATE Optimize its corresponding adversarial samples $\boldsymbol{x}_i^{\prime}$ by PGD attack with $T$ iterations:\\
		\quad $\boldsymbol{x}_i^\prime \gets\mathcal{S}_{\epsilon}(\boldsymbol{x}_i^\prime+\eta\cdot \operatorname{sign}(\nabla_{\boldsymbol{x}_i^\prime}\mathcal{L}_{adv}(\boldsymbol{x}_i^\prime, \boldsymbol{b}_{mi}))) \quad \forall~i;$
		\STATE Update $\theta$ with the gradient descent:\\
		$\theta \gets \theta-\zeta\nabla_{\theta} \frac{1}{n}\sum_{i=1}^n\mathcal{L}_{at}(\boldsymbol{x}_i, \boldsymbol{x}_i^\prime, \boldsymbol{b}_{mi}, \theta);$
		\ENDFOR
		\ENDFOR
		\ENSURE Network parameter $\theta$.
	\end{algorithmic}
\end{algorithm}

Conspicuously, we adopt alternating scheme to optimize the hashing network. Firstly, we maximize $\mathcal{L}_{at}$ with fixed network parameter $\theta$ to generate adversarial examples. Then we optimize the hashing network over $\theta$ to enhance its robustness against adversarial examples as well as maintain its performance on clean samples by minimizing $\mathcal{L}_{at}$. The two steps are iteratively optimized until the hashing network converges to local optima. The overall optimization process of SAAT is outlined in Algorithm \ref{alg:saat}.

\begin{table*}[ht]
\small
\begin{center}
\caption{\small Results of different attack methods on three datasets. We evaluate the attack performance with MAP(\%) criteria for non-targeted attacks (4-th to 6-th rows). For targeted attacks, we use the t-MAP(\%) values to show the results (7-th to 11th rows). The `Original' in table is to query with benign samples, where the MAP values denote the retrieval performance of hashing model without attack.}
\label{tab:map}
\resizebox{0.98\textwidth}{!}{
\begin{tabular}{lr|cccc|cccc|cccc}
\hline
\multirow{2}{*}{Method} & \multirow{2}{*}{Metric} & \multicolumn{4}{c|}{FLICKR-25K}& \multicolumn{4}{c|}{NUS-WIDE}& \multicolumn{4}{c}{MS-COCO} \\
\cline{3-14}
&  & 16bits & 32bits & 48bits & 64bits & 16bits & 32bits & 48bits & 64bits&  16bits & 32bits & 48bits & 64bits\\
\hline
Original & MAP &  81.33 & 82.28 & 82.47 & 81.85 & 76.70 & 77.47 & 77.74 & 78.11 & 56.26 & 57.41 & 56.70 & 56.64 \\
\hline
HAG & MAP & 22.53 & 21.68 & 21.96 & 22.56 & 13.62 & 13.32 & 13.67 & 13.55 & 12.57 & 13.97 & 13.90 & 13.88 \\
SDHA & MAP & 21.82 & 19.37 & 19.39 & 19.20 & 13.85 & 12.98 & 14.49 & 14.56 & 11.61 & 12.38 & 12.63 & 13.08 \\
SAAT (Ours) & MAP & \bf{14.92} & \bf{14.53} & \bf{14.59} & \bf{14.82} & \bf{11.93} & \bf{11.89} & \bf{11.85} & \bf{12.02} & \bf{9.96} & \bf{11.21} & \bf{10.96} & \bf{10.87} \\
\hline
P2P & t-MAP & 81.45 & 81.83 & 82.49 & 82.98 & 66.94 & 69.19 & 69.20 & 69.12 & 51.96 & 52.74 & 52.62 & 51.73 \\
DHTA & t-MAP & 82.98 & 83.51 & 83.67 & 83.94 & 69.10 & 70.92 & 71.03 & 70.93 & 52.57 & 53.38 & 53.23 & 52.34 \\
ProS-GAN & t-MAP & 73.06 & 66.44 & 62.34 & 87.15 & \bf{72.33} & \bf{74.29} & 73.96 & 72.30 & 33.02 & 33.30 & 31.63 & 31.28 \\
THA & t-MAP & 86.80 & 87.84 & 86.94 & 86.73 & 69.78 & 72.98 & 73.90 & 70.29 & 53.18 & 52.88 & 47.80 & 37.63 \\
SAAT (Ours) & t-MAP & \bf{88.62} & \bf{88.90} & \bf{89.07} & \bf{89.00} & 70.68 & 73.99 & \bf{74.00} & \bf{74.33} & \bf{55.75} & \bf{58.00} & \bf{56.97} & \bf{56.58} \\
\hline
\end{tabular}
}
\end{center}
\vspace{-0.5cm}
\end{table*}

\begin{figure*}[ht]
\centering 
\subfigure[Non-targeted]{ 
    \label{fig:pr:non-targeted}  
    \begin{minipage}[b]{0.48\textwidth} 
        \centering
        \includegraphics[width=0.49\textwidth]{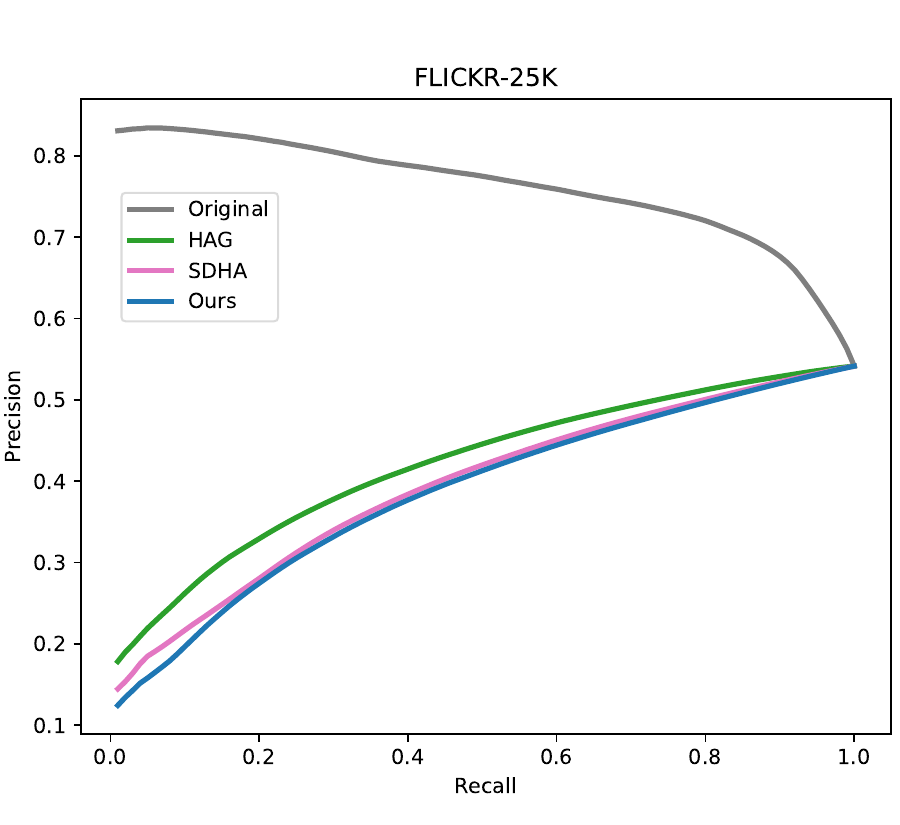}
        \includegraphics[width=0.49\textwidth]{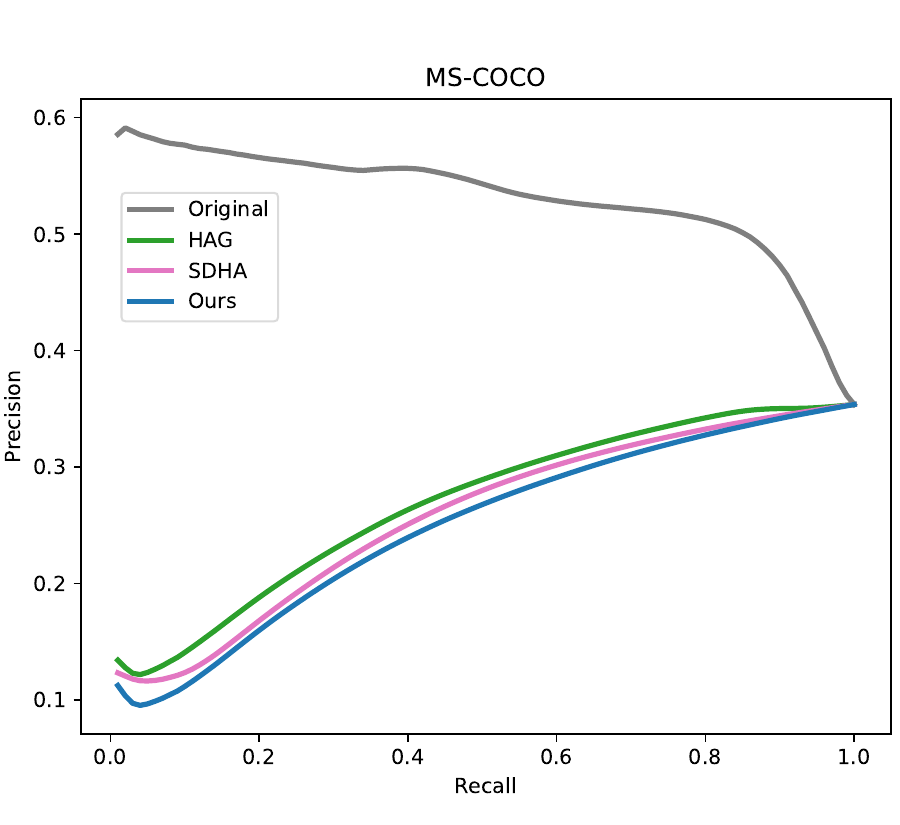}
    \end{minipage}
    }
\subfigure[Targeted]{
    \label{fig:pr:targeted}  
    \begin{minipage}[b]{0.48\textwidth} 
        \centering
        \includegraphics[width=0.49\textwidth]{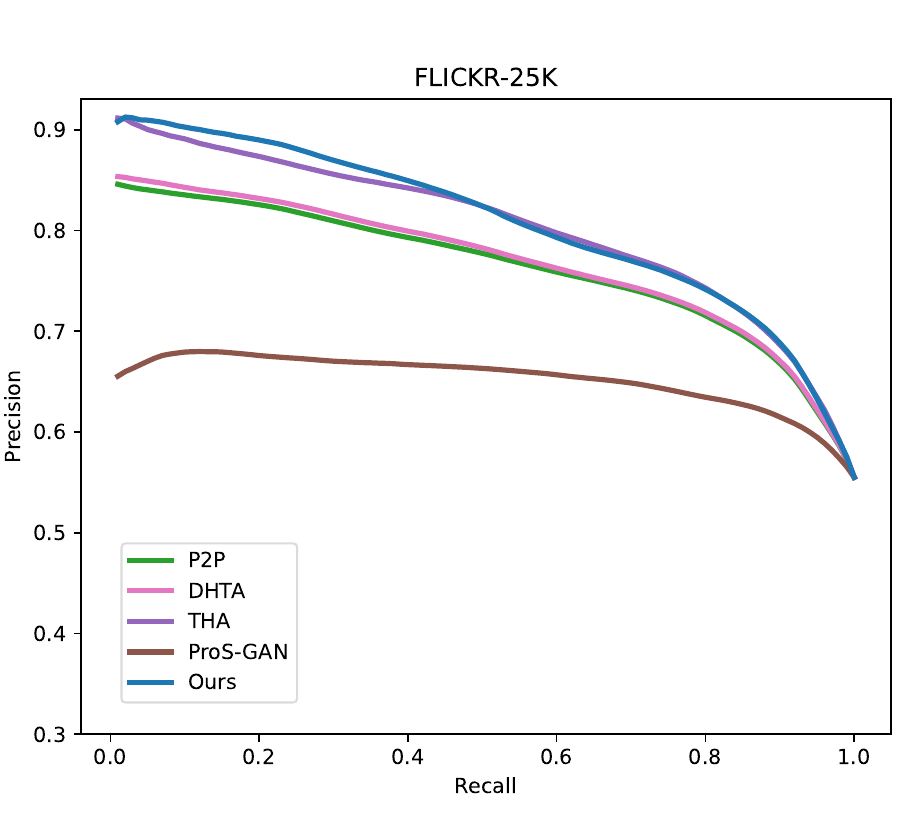}
        \includegraphics[width=0.49\textwidth]{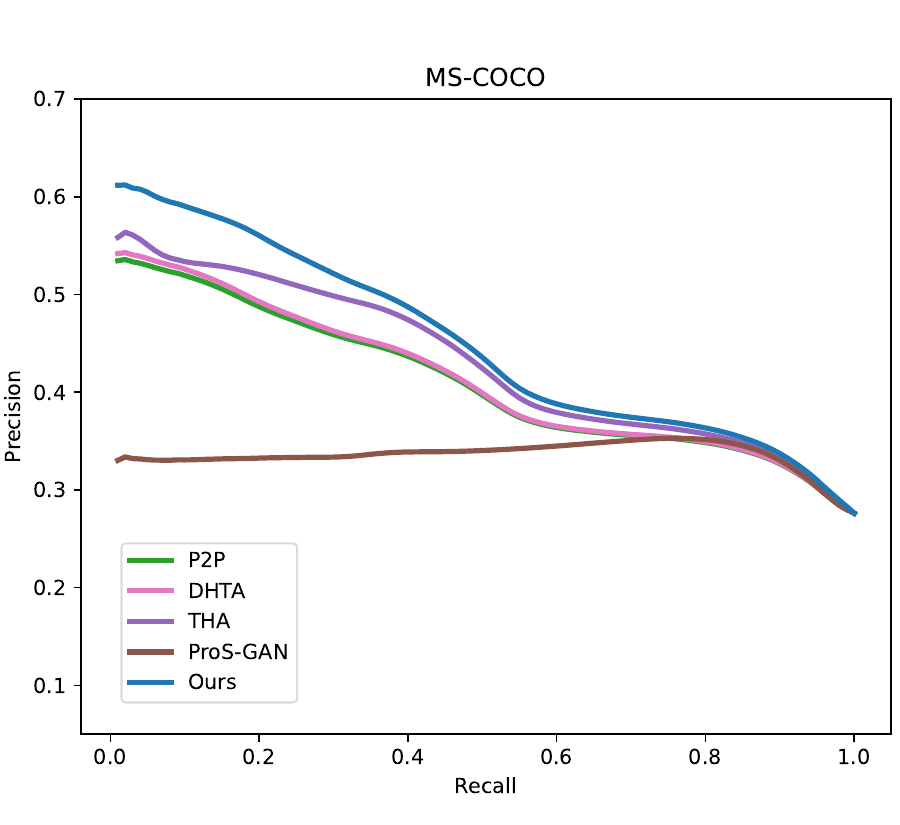}
    \end{minipage}
    }
\caption{\small Precision-Recall curves on FLICKR-25K and MS-COCO under 32 bits code length.}
\label{fig:pr}
\end{figure*}

\subsection{Discussion on the Difference from the Related Works}
Regarding seeking optimal representatives, our work has some similarities with ProS-GAN \cite{wang2021prototype} and THA \cite{wang2021targeted}. However, there are some fundamental differences between our approach and theirs. 1) First, the optimization goal is different. ProS-GAN and THA are designed for targeted adversarial attacks, while our proposed mainstay codes are adaptive to both non-targeted and targeted attacks. Importantly, we provide the first fundamental minimax-form adversarial training for deep hashing with the guidance of mainstay codes. 2) Moreover, ProS-GAN and THA adopt a neural network to learn the prototype code of the target label, which is a parameter-sensitive scheme and results in a lack of theoretical guarantees for the quality of generated representatives. In contrast, we present a discriminative mainstay features learning (DMFL) with a provable mathematical formula to obtain the mainstay codes with global discriminative superiority.

Moreover, the generation of mainstay codes is similar to contrastive learning \cite{chen2020simple,khosla2020supervised} but with three differences. 1) First, the purpose is different. Typically, contrastive learning desires to optimize a network with good generalization capabilities via a self-supervised approach. Differently, we aim to generate a representative semantic feature for each input sample to supervise the adversarial learning of deep hashing. 2) Secondly, it is different for the optimization process. Contrastive learning involves repeatedly measuring the similarity between input and multiple positive and negative samples to adjust the network parameters through numerous iterations. In contrast, we can instantly calculate the mainstay code of the input sample by exploiting the proposed DMFL, dramatically reducing the computational overhead. 3) Furthermore, contrastive learning occurs in the continuous feature space, while mainstay code is produced in the discrete binary space. Hence, solving the mainstay code by gradient descent is intractable, yet our proposed DMFL cleverly and efficiently addresses this problem.

\section{Experiments} \label{sec:experiments}
\subsection{Experimental Setup}
\subsubsection{Datasets.}
We conduct extensive experiments to evaluate our methods on three well-known datasets: \textbf{FLICKR-25K} \cite{huiskes2008mir}, \textbf{NUS-WIDE} \cite{chua2009nus} and \textbf{MS-COCO} \cite{lin2014microsoft}. \textbf{FLICKR-25K} consists of 25,000 Flickr images annotated with 38 labels. We pick 1000 instances from the whole dataset as queries, while the remaining are regarded as the database. Moreover, we randomly select 5,000 images from the database for training hashing models and producing mainstay codes. \textbf{NUS-WIDE} has 269,648 samples with 81 concepts. Following \cite{wang2021prototype}, we select a subset containing 21 most frequent concepts with 193,734 images as database and 2,100 images for testing. Besides, we select 10,500 images from the database for training. \textbf{MS-COCO} involves 123,287 samples after combining the training and validation sets, where each sample is labeled with 80 classes. Following \cite{cao2017hashnet}, we randomly select 5,000 images as queries and the rest as the database. For the training set, 10,000 images are randomly picked from the database.

\subsubsection{Baselines.}
Following \cite{yang2018adversarial}, we select DPH \cite{yang2018adversarial} as target hashing model to be attacked, which is a generic deep hashing retrieval model. Particularly, we can change it to any other deep hashing models. AlexNet \cite{krizhevsky2012imagenet} and VGGs \cite{simonyan2014very} are chosen as the backbone networks to implement DPH. Specifically, we replace their last classifier layer with a hashing layer, which involves a fully connected layer with $K$ hidden units and a tanh activation. We also evaluate the generalizability of our method on other hashing methods, including DPSH \cite{li2016feature}, HashNet \cite{cao2017hashnet} and CSQ \cite{yuan2020central}. For estimating the effectiveness of our methods, we implement previous attack and defense methods in hashing retrieval, covering two non-targeted attacks (\textit{i.e.}, HAG \cite{yang2018adversarial} and SDHA \cite{lu2021smart}), three targeted attack methods (\textit{i.e.}, P2P \cite{bai2020targeted}, DHTA \cite{bai2020targeted}, ProS-GAN \cite{wang2021prototype} and THA \cite{wang2021targeted}), and the only one defense algorithm ATRDH \cite{wang2021targeted}. For targeted attacks, we randomly choose a label for attack, which does not share the same class with the original label. To make fair comparisons, the detailed implementations of these methods are consistent with the original papers with released codes.

\begin{figure*}[ht]
\centering 
\subfigure[Non-targeted]{ 
    \label{fig:topn:non-targeted}  
    \begin{minipage}[b]{0.48\textwidth} 
        \centering
        \includegraphics[width=0.49\textwidth]{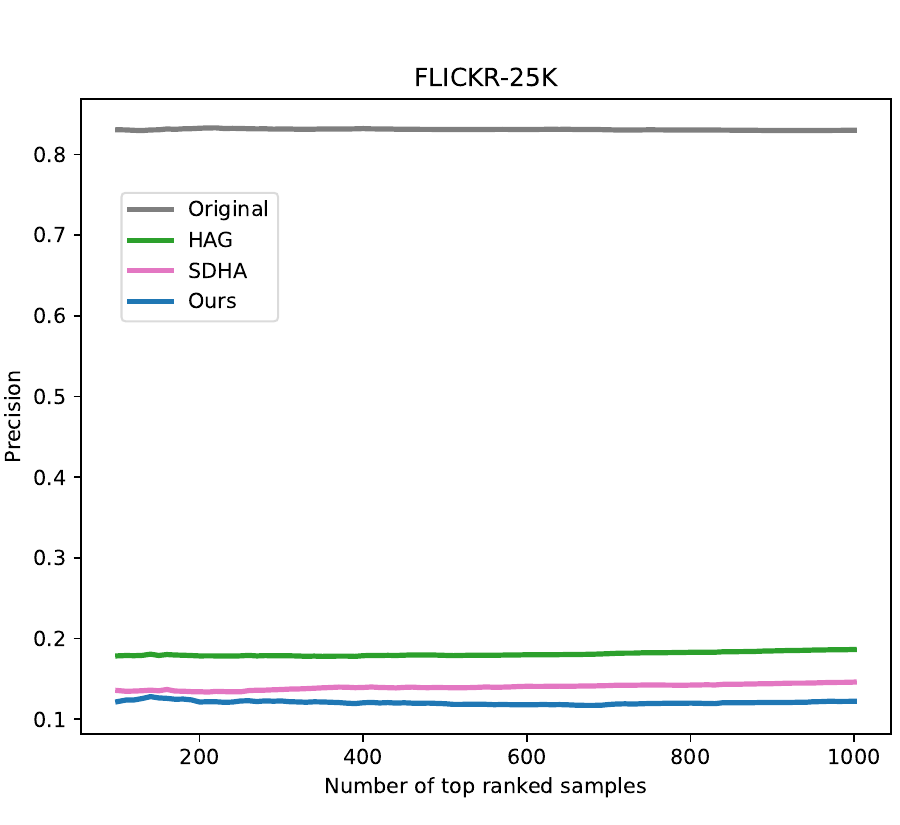}
        \includegraphics[width=0.49\textwidth]{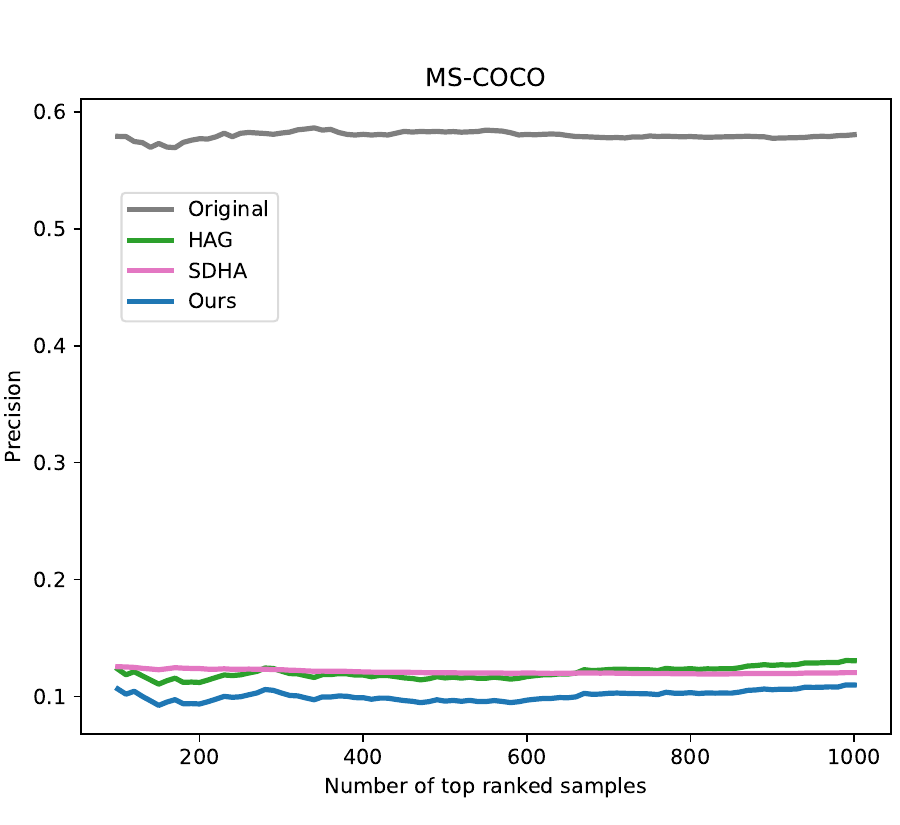}
    \end{minipage}
    }
\subfigure[Targeted]{
    \label{fig:topn:targeted}  
    \begin{minipage}[b]{0.48\textwidth} 
        \centering
        \includegraphics[width=0.49\textwidth]{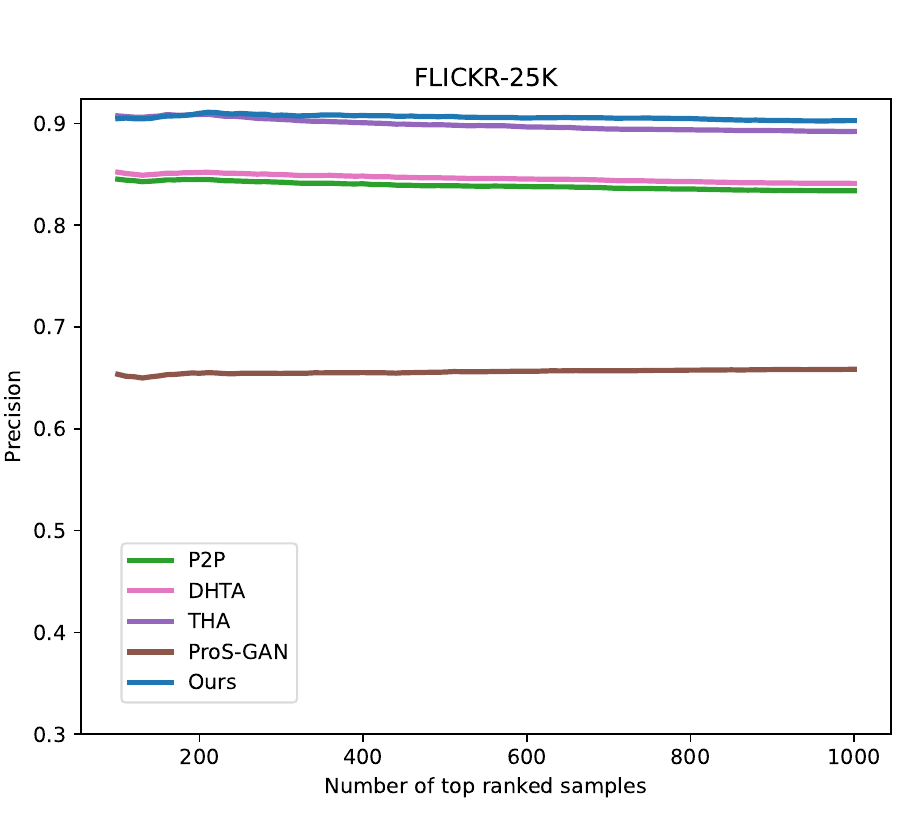}
        \includegraphics[width=0.49\textwidth]{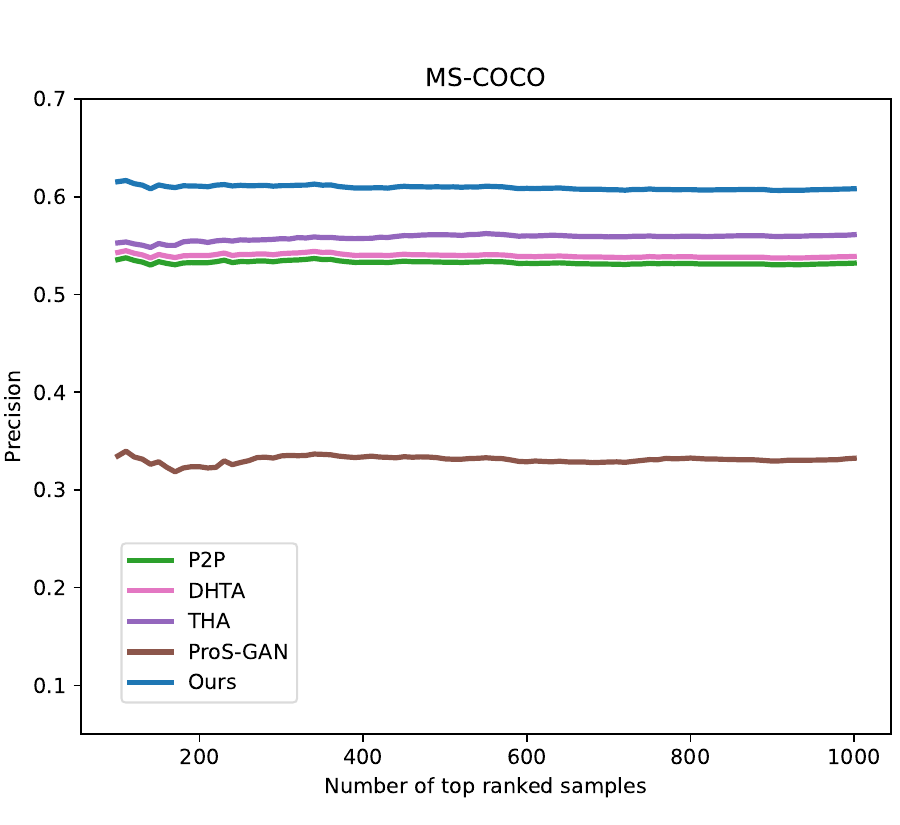}
    \end{minipage}
    }
\caption{\small Precision@1000 curves on FLICKR-25K and MS-COCO under 32 bits code length.} 
\label{fig:topn}
\vspace{-3ex}
\end{figure*}

\begin{figure*}[ht]
\subfigure[Non-targeted]{
    \begin{minipage}[b]{0.48\textwidth} 
        \centering
        \includegraphics[width=\textwidth]{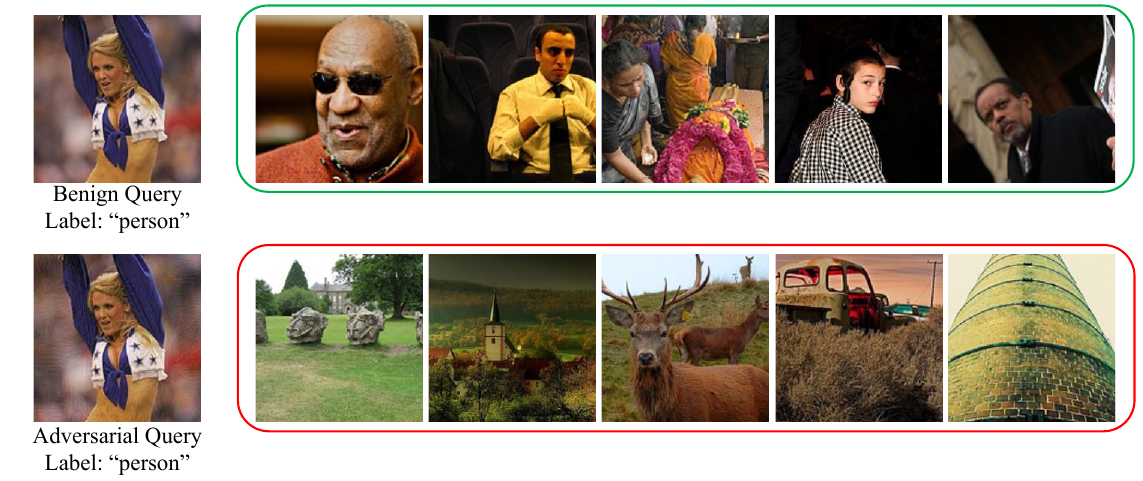}
    \end{minipage}
}
\subfigure[Targeted]{
    \begin{minipage}[b]{0.48\textwidth} 
        \centering
        \includegraphics[width=\textwidth]{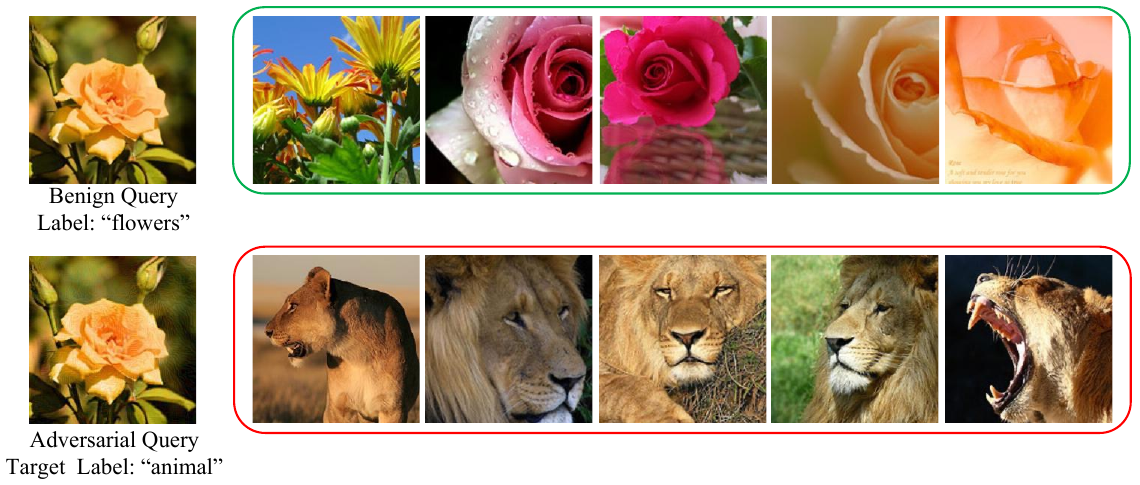}
    \end{minipage}
}
\caption{\small Retrieval examples on NUS-WIDE with the benign query and its adversarial version. We provide examples of non-targeted attacks and targeted attacks in (a) and (b), respectively. For each example, the two boxes represent the top-5 retrieved images of the natural and adversarial queries, respectively.}
\label{fig:retrieval_examples}
\vspace{-2ex}
\end{figure*}

\subsubsection{Implementation Details.} 
We utilize stochastic gradient descent (SGD) \cite{robbins1951stochastic} with a learning rate $0.01$ and momentum $0.9$ as an optimizer to pre-train the target hashing models. All images are resized to $224 \times 224$ and normalized in $[0,1]$ before feeding in hashing models.
For the proposed attack algorithm, we apply PGD \cite{bai2020targeted} to optimize adversarial samples. The step size $\eta$ and the number of iterations $T$ are set to $1/255$ and $100$, respectively. The perturbation budget $\epsilon$ is fixed to $8/255$. Similar to \cite{yang2018adversarial}, we set the hyper-parameter $\alpha$ with $0.1$ during the ﬁrst $50$ iterations, then update it every $10$ iterations according to $[0.2, 0.3, 0.5, 0.7, 1.0]$ during the final 50 iterations.
For our defense method SAAT, we conduct 20 epochs for adversarial training. During the training process, we empirically set $\eta$ and $T$ in PGD with 2 and 7, respectively, to generate adversarial samples in each epoch. The weighting factors $\lambda$ and $\mu$ of Eq. (\ref{eq:obj_at}) are set as $1$ and $10^{-4}$, respectively.

\subsubsection{Protocols.}
Following \cite{yang2018adversarial}, we adopt MAP (mean average precision) to evaluate the performance of non-targeted attacks and calculate MAP values on the top 5000 results from the database. Specifically, for targeted attacks, we employ t-MAP (targeted mean average precision) \cite{bai2020targeted} to appraise their results. As t-MAP takes the target labels as the test labels, the higher the t-MAP, the stronger the targeted attack. In addition, we also provide Precision-Recall (PR) curves and precision@topN curves for comprehensive analysis.

\subsection{Adversarial Attack Results}
As shown in Table \ref{tab:map}, we present the detailed MAP scores of non-targeted attacks on three benchmarks. A lower MAP means a stronger performance of non-targeted attacks. The \textit{Original} in Table \ref{tab:map} is a query using original samples without any additive noise, where the corresponding MAP reflects the retrieval performance of the attack-free hashing model. 
From the results, we can observe that our method outperforms all non-targeted attacks and significantly drops the MAP values on three benchmarks with the hash bits varying from 16 to 64. Compared with SDHA \cite{lu2021smart}, the best non-targeted attack method, the MAP scores of our method decline an average of 2.98\% for all cases. Especially, on the FLICKR-25K, our method outperforms SDHA by over 4.38\% for any bits. The superior behavior of our method owes to the high-quality mainstay code, which represents the globally optimal semantic of a given sample by preserving the similarity with positives and irrelevancy with negatives simultaneously. In contrast, HAG and SDHA just use the information from benign samples and positive samples, respectively. It is worth noting that our attack merely runs 100 iterations during optimization, but other methods use up to 1,500 iterations, or even 2,000 iterations, which further reveals the efficiency of our method.

Results in terms of t-MAP for targeted attacks are also given in Table \ref{tab:map}. It can be observed that our method achieves the best performance on targeted attacks in most cases. When comparing with the state-of-the-art targeted attack, THA \cite{wang2021targeted}, our attack accomplishes average boosts of 1.82\%, 1.52\% and 8.95\% for different bits on FLICKR-25K, NUS-WIDE, and MS-COCO, respectively. Specifically, although ProS-GAN yields better results for 16bits and 32bits on NUS-WIDE, it does not work well on other datasets, especially MS-COCO. This is because the generative framework of ProS-GAN is parameter-sensitive and lacks generalization on different datasets. Moreover, the prototype network they designed is hard to fit multi-label cases (\textit{e.g.}, MS-COCO with 80 classes). Differently, our mainstay codes are obtained by discriminative mainstay feature learning, which can be well adapted to various datasets.

\begin{table*}[htbp]
\small
\begin{center}
\caption{\small MAP (\%) of non-targeted attack methods after adversarial training by ATRDH and our SAAT.}
\label{tab:defense_nontargeted}
\resizebox{0.98\textwidth}{!}{
\begin{tabular}{ll|cccc|cccc|cccc}
\hline
\multirow{2}{*}{Defense} & \multirow{2}{*}{Attack} & \multicolumn{4}{c|}{FLICKR-25K}& \multicolumn{4}{c|}{NUS-WIDE}& \multicolumn{4}{c}{MS-COCO} \\
\cline{3-14}
& & 16bits & 32bits & 48bits & 64bits & 16bits & 32bits & 48bits & 64bits&  16bits & 32bits & 48bits & 64bits\\
\hline
\multirow{4}{*}{ATRDH} & Original & 71.35 & 72.16 & 72.29 & 72.08 & 64.58 & 64.57 & 67.20 & 67.90 & 49.41 & 50.89 & 50.38 & 51.57  \\
~ & HAG & 41.92 & 42.36 & 43.80 & 44.08 & 41.59 & 41.93 & 42.31 & 42.29 & 27.19 & 26.78 & 26.20 & 26.75 \\
~ & SDHA  & 39.09 & 37.77 & 38.39 & 38.60 & 41.61 & 40.36 & 40.78 & 40.59 & 28.09 & 27.17 & 27.55 & 27.75 \\
~ & Ours  & \bf{32.68} & \bf{32.40} & \bf{33.39} & \bf{33.32} & \bf{38.53} & \bf{38.16} & \bf{38.95} & \bf{38.99} & \bf{23.93} & \bf{23.24} & \bf{23.14} & \bf{22.82} \\
\hline
\multirow{4}{*}{SAAT(Ours)} & Original & 74.19 & 73.49 & 73.41 & 70.07 & 61.05 & 60.51 & 61.06 & 60.57 & 46.91 & 49.41 & 50.14 & 52.01 \\
~ & HAG & 38.05 & 43.76 & 48.22 & 60.89 & 53.07 & 52.72 & 52.56 & 52.90 & 33.83 & 34.01 & 35.77 & 36.69 \\
~ & SDHA & 35.26 & 40.52 & 45.78 & 59.62 & 53.22 & 52.15 & 52.57 & 52.83 & 35.99 & 35.07 & 37.23 & 36.91 \\
~ & Ours & \bf{30.55} & \bf{34.77} & \bf{39.35} & \bf{53.76} & \bf{50.35} & \bf{50.33} & \bf{50.20} & \bf{50.16} & \bf{30.98} & \bf{30.87} & \bf{32.63} & \bf{33.43} \\
\hline
\end{tabular}
}
\end{center}
\end{table*}

\begin{table*}[htbp]
\small
\begin{center}
\caption{\small t-MAP (\%) of targeted attack methods after adversarial training by ATRDH and our SAAT. The `Original' in table indicates the MAP value of hashing model without attack.}
\label{tab:defense_targeted}
\resizebox{0.98\textwidth}{!}{
\begin{tabular}{ll|cccc|cccc|cccc}
\hline
\multirow{2}{*}{Defense} & \multirow{2}{*}{Attack} & \multicolumn{4}{c|}{FLICKR-25K}& \multicolumn{4}{c|}{NUS-WIDE}& \multicolumn{4}{c}{MS-COCO} \\
\cline{3-14}
& & 16bits & 32bits & 48bits & 64bits & 16bits & 32bits & 48bits & 64bits&  16bits & 32bits & 48bits & 64bits\\
\hline
\multirow{6}{*}{ATRDH} & Original & 71.81 & 72.40 & 72.99 & 72.26 & 65.60 & 66.96 & 67.04 & 68.15 & 49.71 & 50.67 & 51.25 & 50.54 \\
~ & P2P & 72.54 & 71.76 & 72.05 & 71.81 & 46.01 & 48.80 & 49.27 & 49.52 & 35.69 & 37.06 & 37.67 & 37.75 \\
~ & DHTA  & 73.45 & 72.97 & 72.90 & 73.56 & 47.53 & 50.39 & 50.76 & 51.37 & 36.13 & 37.54 & 38.14 & 37.88 \\
~ & ProS-GAN  & 49.76 & 49.46 & 48.64 & 49.71 & 29.82 & 28.92 & 28.67 & 29.25 & 23.34 & 23.73 & 23.22 & 23.29 \\
~ & THA  & 77.37 & 77.97 & 78.07 & 78.32 & \bf{51.29} & 54.85 & 56.00 & 55.46 & 41.13 & 43.31 & 43.35 & 40.92 \\
~ & Ours  & \bf{78.67} & \bf{79.00} & \bf{79.52} & \bf{79.40} & 51.09 & \bf{55.31} & \bf{56.28} & \bf{55.85} & \bf{42.18} & \bf{44.11} & \bf{44.92} & \bf{44.89} \\
\hline
\multirow{6}{*}{SAAT(Ours)} & Original & 73.45 & 72.89 & 72.86 & 71.77 & 61.37 & 61.92 & 61.99 & 61.02 & 49.34 & 50.68 & 52.18 & 53.27 \\
~ & P2P & 73.17 & 71.11 & 69.39 & 65.83 & 40.74 & 37.68 & 37.32 & 38.41 & 33.08 & 33.12 & 33.94 & 34.06 \\
~ & DHTA  & 74.58 & 72.33 & 70.74 & 66.89 & 41.35 & 38.19 & 37.89 & 38.69 & 33.28 & 33.33 & 34.15 & 34.14 \\
~ & ProS-GAN  & 48.24 & 48.43 & 49.20 & 50.35 & 37.99 & 33.81 & 32.93 & 30.71 & 26.61 & 25.29 & 26.09 & 25.09 \\
~ & THA  & 76.47 & 74.38 & 72.27 & 66.33 & 41.28 & 38.71 & 37.83 & 38.02 & 34.27 & 34.37 & 35.57 & 34.39 \\
~ & Ours & \bf{79.65} & \bf{76.51} & \bf{75.07} & \bf{69.97} & \bf{42.02} & \bf{39.40} & \bf{38.72} & \bf{39.67} & \bf{36.72} & \bf{36.28} & \bf{36.97} & \bf{37.76} \\
\hline
\end{tabular}
}
\end{center}
\end{table*}

For a more comprehensive comparison, the retrieval performance on benchmarks in terms of PR curves and precision@topN curves are shown in Fig. \ref{fig:pr} and Fig. \ref{fig:topn}, respectively. We only provide the results on FLICKR-25K and MS-COCO of different methods with 32 bits length. For non-targeted attacks, we can see that the curves of our method are always below all other curves while the opposite is the case in targeted attacks, which demonstrates that our method is able to attack hashing models more effectively. Furthermore, we also provide two examples of the retrieval results on NUS-WIDE with a benign image and its adversarial version generated by our method in Fig. \ref{fig:retrieval_examples} for intuitive understanding.

\subsection{Adversarial Defense Results}
To improve the adversarial robustness of deep hashing networks, we perform the proposed formalized adversarial training algorithm on pre-trained deep hashing models. After the adversarial training, we re-attack these models and the results in terms of MAP are reported in Table \ref{tab:defense_nontargeted}. By comparing Table \ref{tab:defense_nontargeted} with Table \ref{tab:map}, we observe that the MAP values of different attack methods are much higher than no adversarial training, though the original MAP values decrease slightly. For example, under the non-targeted attack SDHA, SAAT brings an average increase of approximately 25.35\%, 38.72\% and 23.87\% on FLICKR-25K, NUS-WIDE and MS-COCO, respectively. For our proposed semantic-aware attack, SAAT improves by an average of about 24.94\%, 38.33\%, and 21.22\%. To further verify the effectiveness of our well-designed SAAT, we compare SAAT with ATRDH \cite{wang2021targeted} under the same experiment settings and the results of ATRDH are also illustrated in Table \ref{tab:defense_nontargeted}. As we can see, the original MAP values of SAAT and ATRDH are close, but our SAAT achieves a signiﬁcant performance boost in terms of the MAP under various attacks. For example, with respect to our proposed attack algorithm, SAAT exceeds ATRDH by an average of 6.66\%, 11.60\%, 8.69\% for FLICKR-25K, NUS-WIDE, and MS-COCO, respectively. The above phenomena show that SAAT can learn more robust hashing codes than ATRDH, because SAAT is a standard minimax-based adversarial training.

\begin{table*}[ht]
\small
\begin{center}
\caption{\small Theoretical Values of different attack methods. For each method, we utilize the representative codes of them to calculate theoretical MAP(t-MAP).}
\label{tab:theory-map}
\resizebox{0.98\textwidth}{!}{
\begin{tabular}{lr|cccc|cccc|cccc}
\hline
\multirow{2}{*}{Method} & \multirow{2}{*}{Metric} & \multicolumn{4}{c|}{FLICKR-25K}& \multicolumn{4}{c|}{NUS-WIDE}& \multicolumn{4}{c}{MS-COCO} \\
\cline{3-14}
& & 16bits & 32bits & 48bits & 64bits & 16bits & 32bits & 48bits & 64bits&  16bits & 32bits & 48bits & 64bits\\
\hline
Original & MAP &  81.33 & 82.28 & 82.47 & 81.85 & 76.70 & 77.47 & 77.74 & 78.11 & 56.26 & 57.41 & 56.70 & 56.64 \\
\hline
HAG & MAP & 22.67 & 21.54 & 21.73 & 22.19 & 13.51 & 13.32 & 13.65 & 13.54 & 13.14 & 14.35 & 14.15 & 13.83 \\
SDHA & MAP & 18.89 & 18.55 & 18.65 & 18.80 & 14.13 & 13.82 & 13.99 & 14.17 & 9.99 & 11.60 & 11.19 & 11.37 \\
SAAT(Ours) & MAP & \bf{14.90} & \bf{14.51} & \bf{14.57} & \bf{14.72} & \bf{11.69} & \bf{11.77} & \bf{11.77} & \bf{11.75} & \bf{9.76} & \bf{11.09} & \bf{10.76} & \bf{10.68} \\
\hline
P2P & t-MAP & 81.35 & 81.87 & 82.60 & 83.10 & 66.87 & 69.14 & 69.11 & 69.11 & 52.23 & 52.89 & 52.90 & 52.37 \\
DHTA & t-MAP & 82.93 & 83.56 & 83.72 & 84.13 & 69.13 & 70.85 & 70.82 & 70.75 & 52.82 & 53.57 & 53.54 & 52.93 \\
ProS-GAN & t-MAP & 86.47 & 86.88 & 87.05 & 59.45 & 70.70 & 73.44 & 73.48 & 72.93 & 50.85 & 36.19 & 40.60 & 35.85 \\
THA & t-MAP & 86.67 & 87.86 & 87.36 & 86.81 & 70.18 & 73.20 & 73.73 & 67.25 & 51.66 & 49.74 & 50.61 & 44.72 \\
SAAT(Ours) & t-MAP & \bf{88.63} & \bf{88.91} & \bf{89.11} & \bf{89.04} & \bf{70.92} & \bf{73.97} & \bf{74.10} & \bf{74.35} & \bf{59.16} & \bf{60.14} & \bf{59.63} & \bf{58.90} \\
\hline
\end{tabular}
}
\end{center}
\vspace{-0.3cm}
\end{table*}

\begin{table}[ht]
\small
\centering
\caption{MAP (\%) of non-targeted attack with different iterations on NUS-WIDE.}
\label{tab:iteration-map}
\resizebox{0.35\textwidth}{!}{
\begin{tabular}{c|cccc}
\hline
Iterations & 16bits & 32bits & 48bits & 64bits\\
\hline
0 & 76.70 & 77.47 & 77.74 & 78.11 \\
1 & 62.82 & 63.90 & 63.71 & 63.77 \\
10 & 13.18 & 13.47 & 13.45 & 13.44 \\
50 & 12.04 & 12.06 & 12.13 & 11.96 \\
100 & 11.96 & 11.97 & 12.12 & 11.93 \\
500 & 11.80 & 11.90 & 12.10 & 11.96 \\
1000 & \bf{11.76} & \bf{11.88} & \bf{12.09} & \bf{11.88} \\
\hline
Theory & 11.69 & 11.77 & 11.77 & 11.75 \\
\hline
\end{tabular}
}
\end{table}

A similar situation can be observed in the targeted attack shown in Table \ref{tab:defense_targeted}. In the case of THA, SAAT achieves an average improvement of over 14.71\%, 32.77\% and 13.22\% for different bits on FLICKR-25K, NUS-WIDE and MS-COCO, respectively, in comparison to the models without adversarial training. Additionally, compared with ATRDH, SAAT outscores ATRDH by an average of 5.57\%, 15.44\%, 7.52\% on FLICKR-25K, NUS-WIDE, and MS-COCO, respectively. These cases all confirm that the proposed adversarial training strategy can effectively improve the robustness of deep hashing models against both targeted and non-targeted attacks. It is worth remarking that the attack performance of ProS-GAN on the hashing models retrained by SAAT dramatically decreases, which indicates that our defense method can basically resist the attack of ProS-GAN. This behavior is more pronounced on ATRDH, which may be attributed to the fact that adversarial training is more effective for generative-based attack methods.

Besides, it can be seen from the Table \ref{tab:defense_nontargeted} and \ref{tab:defense_targeted} that our attack achieves the-sate-of-the-art attack performance on all defense models, which further validates the superiority of our attack algorithm.

\begin{figure}[!t]
\subfigure[$\lambda$]{ 
    \centering
    \label{fig:ablation:lambda}
    \includegraphics[width=0.23\textwidth]{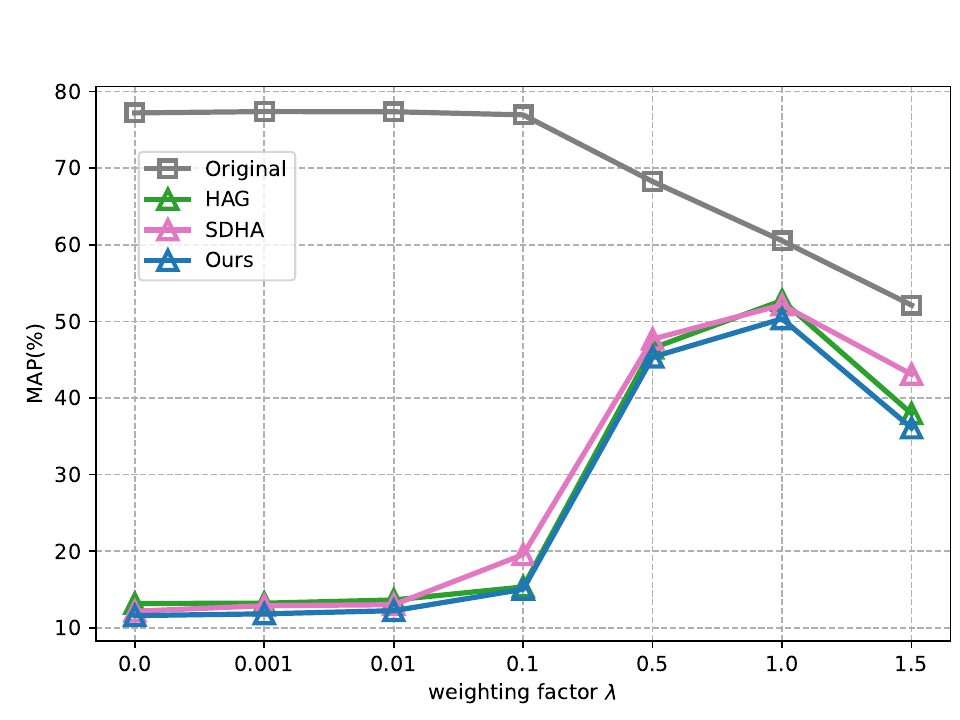}}
\subfigure[$\mu$]{
    \centering
    \label{fig:ablation:mu}
    \includegraphics[width=0.23\textwidth]{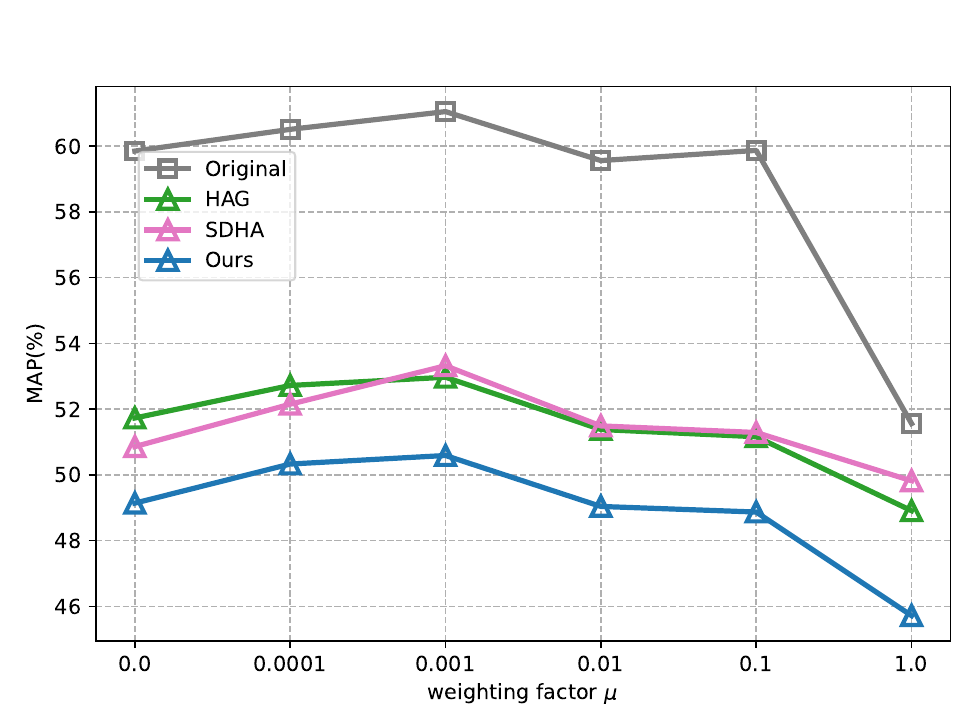}}
\caption{\small MAP (\%) on NUS-WIDE for our adversarial training with different $\lambda$ and $\mu$.}
\label{fig:ablation} 
\end{figure}

\begin{table}[!t]
\small
\centering
\caption{Perceptibility of adversarial perturbations under non-targeted attacks. $L_{\infty}$ and $L_2$ are matrix norms of adversarial perturbations, and MSE is mean squared error between benign samples and its adversarial versions. All the results are averaged over the entire test set.}
\label{tab:perceptibility}
\resizebox{0.48\textwidth}{!}{
\begin{tabular}{c|ccc|ccc|ccc}
\hline
\multirow{2}{*}{Metric} & \multicolumn{3}{c|}{FLICKR-25K}& \multicolumn{3}{c|}{NUS-WIDE}& \multicolumn{3}{c}{MS-COCO} \\
\cline{2-10}
 & HAG & SDHA & Ours & HAG & SDHA & Ours & HAG & SDHA & Ours \\
\hline
$L_{\infty}$ & 6.26 & 6.03 & \bf{5.27} & 6.35 & 6.13 & \bf{5.61} & 6.23 & 6.66 & \bf{5.39} \\
$L_2$ & 1.24 & 1.17 & \bf{1.04} & 1.25 & 1.19 & \bf{1.08} & 1.21 & 1.23 & \bf{1.01} \\
MSE($\times 10^{-4}$) & 7.18 & 6.73 & \bf{5.20} & 7.29 & 6.89 & \bf{5.68} & 7.08 & 8.26 & \bf{5.44} \\
\hline
MAP(\%) & 21.94 & 17.66 & \bf{14.97} & 13.20 & 12.94 & \bf{11.96} & 13.57 & 12.39 & \bf{11.12} \\
\hline
\end{tabular}
}
\end{table}

\subsection{Analysis and Discussions}
\subsubsection{Attack Results in Theory}
Besides actual attack performance, we also record the MAP(t-MAP) values in theory of various attack methods to compare the theoretical upper bounds that they can reach, which is shown in Table \ref{tab:theory-map}. Intuitively, we calculate the theoretical MAP by directly retrieving the contents in Hamming space with representative codes of each method instead of the hash codes corresponding to the adversarial samples. Such as the mainstay code of our method, the anchor code of DHTA, and the prototype code of ProS-GAN. Obviously, our attack outperforms all comparison methods on both targeted and non-targeted attacks, which further verifies the superiority of our well-designed mainstay code. Making a comparison between Table \ref{tab:map} and Table \ref{tab:theory-map}, we notice that our attack algorithm converges well, as the actual attack performance is considerably approached to the theoretical results.

\subsubsection{Effect of $T$ in PGD}
Table \ref{tab:iteration-map} presents MAP results of adversarial examples under non-targeted attack with different optimizing iterations (\textit{i.e.}, $T$). Adversarial examples at iteration 0 correspond to the benign examples without adversarial noise. As expected, MAP drops quickly with the number of iterations. It is worth noting that MAP reduces over 82\% of the original MAP at 10 iterations in all cases, which indicates our attack method is much more efficient. Continued optimization of adversarial examples can further reduce the MAP values until the 100th iteration. In addition, the MAP values of the 500 and 1000 iterations are almost invariant and close to the theoretical values. These results imply that we have a good balance between the effectiveness and efficiency of adversarial attacks with $T=100$.

\subsubsection{Perceptibility}
\begin{figure*}
\subfigure[FLICKR-25K]{                  
    \label{fig:adversarial:FLICKR-25K}  
    \begin{minipage}[b]{0.343\textwidth} 
        \centering
        \includegraphics[width=\textwidth]{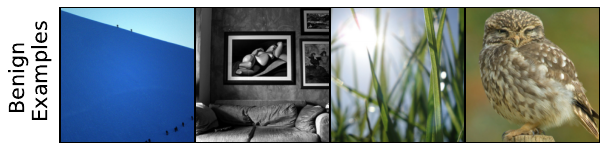}
        \includegraphics[width=\textwidth]{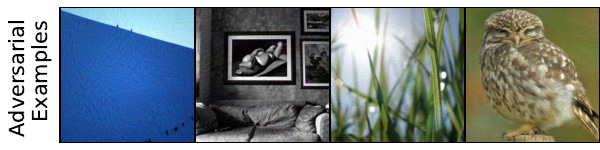}
        \includegraphics[width=\textwidth]{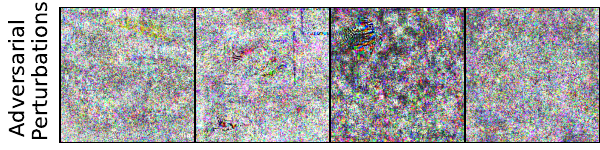}
    \end{minipage}
}
\subfigure[NUS-WIDE]{ 
    \label{fig:adversarial:NUS-WIDE}  
    \begin{minipage}[b]{0.31\textwidth} 
        \centering 
        \includegraphics[width=\textwidth]{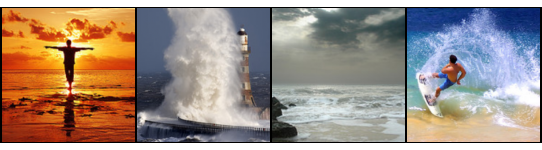}
        \includegraphics[width=\textwidth]{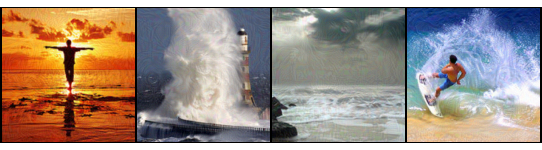}
        \includegraphics[width=\textwidth]{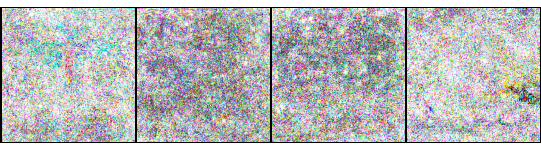}
    \end{minipage}
}
\subfigure[MS-COCO]{ 
    \label{fig:adversarial:MS-COCO}  
    \begin{minipage}[b]{0.31\textwidth} 
        \centering 
        \includegraphics[width=\textwidth]{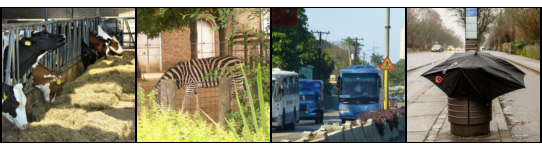}
        \includegraphics[width=\textwidth]{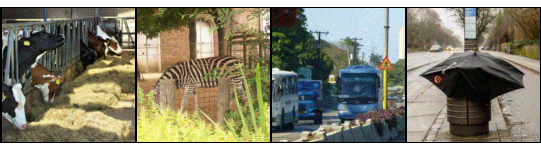}
        \includegraphics[width=\textwidth]{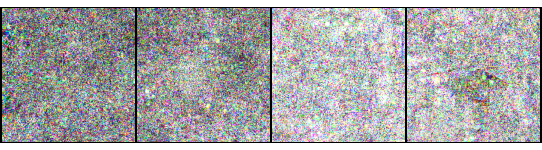}
    \end{minipage}
}
\caption{Visualization of generated adversarial examples. Given benign images (the first row), we produce corresponding adversarial examples using our attack method, as shown in the second row. For better presentation, we show the normalized adversarial perturbation for each sample in the last row.}
\label{fig:visualization} 
\end{figure*}

Beyond attack performance, perceptibility is another essential aspect in evaluating the quality of adversarial examples. Hence, we report the perceptibility of adversarial noise with respect to three metrics, including $L_{\infty}$, $L_2$ and MSE. The results for all datasets are shown in Table \ref{tab:perceptibility}. Since the adversarial perturbation is normally desired to be imperceptible, the higher the perceptibility, the worse the visual quality of the adversarial example. As we can see, all the  results of our method are very small and outperform those of the other two non-targeted attacks by large margins. This demonstrates that our approach achieves excellent attack performance while ensuring the imperceptibility of the learned perturbations. Some visual examples of adversarial images generated by our method in Fig. \ref{fig:visualization} also verify this point.

\begin{table*}[!t]
\small
\begin{center}
\caption{\small MAP (\%) of non-targeted attacks for different hashing models on NUS-WIDE. For each deep hashing algorithm, we implement it with four backbone networks, including AlexNet\cite{krizhevsky2012imagenet} and VGG family \cite{simonyan2014very}.}
\label{tab:univer}
\resizebox{\textwidth}{!}{
\begin{tabular}{cl|cccc|cccc|cccc}
\hline
\multirow{2}{*}{Defense} & \multirow{2}{*}{Attack} & \multicolumn{4}{c|}{DPSH}& \multicolumn{4}{c|}{HashNet}& \multicolumn{4}{c}{CSQ} \\
\cline{3-14}
&  & AlexNet & VGG11 & VGG16 & VGG19 & AlexNet & VGG11 & VGG16 & VGG19 &  AlexNet & VGG11 & VGG16 & VGG19 \\
\hline
\multirow{4}{*}{\makecell[c]{No \\ Defense}} & Original & 80.97 & 82.00 & 81.96 & 81.78 & 79.65 & 81.91 & 81.13 & 80.56 & 79.26 & 80.71 & 81.53 & 0.8160 \\
~ & HAG  & 14.49 & 13.28 & 14.52 & 13.73 & 12.99 & 11.13 & 10.31 & 11.29 & 16.58 & 14.93 & 19.89 & 14.76 \\
~ & SDHA & 13.72 & 11.06 & 14.30 & 14.00 & 9.59 & 10.43 & 10.59 & 10.03 & 12.99 & 8.85 & 8.53 & 7.95 \\
~ & Ours & \bf{11.32} & \bf{9.70} & \bf{10.50} & \bf{11.27} & \bf{8.10} & \bf{6.18} & \bf{6.36} & \bf{7.15} & \bf{7.18} & \bf{5.92} & \bf{6.22} & \bf{7.18}  \\
\hline
\multirow{4}{*}{ATRDH} & Original & 72.41 & 70.52 & 69.65 & 69.02 & 72.79 & 68.31 & 68.43 & 65.16 & 61.75 & 57.62 & 59.21 & 65.31 \\
~ & HAG & 33.62 & 45.13 & 47.44 & 47.26 & 30.83 & 46.63 & 51.20 & 51.60 & 38.70 & 35.60 & 34.80 & 24.85 \\
~ & SDHA & 34.12 & 39.86 & 44.59 & 46.32 & 25.87 & 42.13 & 47.48 & 47.37 & 38.80 & 34.85 & 36.87 & \bf{19.33} \\
~ & Ours & \bf{31.22} & \bf{33.85} & \bf{35.87} & \bf{38.69} & \bf{23.56} & \bf{41.43} & \bf{46.23} & \bf{45.48} & \bf{36.33} & \bf{30.38} & \bf{27.76} & 29.36  \\
\hline
\multirow{4}{*}{\makecell[c]{SAAT \\ (Ours)}} & Original & 74.83 & 76.70 & 75.99 & 75.74 & 76.38 & 68.75 & 69.56 & 70.46 & 72.94 & 66.79 & 67.65 & 74.96 \\
~ & HAG  & 44.22 & 45.45 & 54.10 & 43.65 & 40.33 & 57.02 & 55.54 & 55.17 & 44.44 & 45.36 & 49.85 & 42.19 \\
~ & SDHA & 47.53 & 47.63 & 55.49 & 45.55 & 41.84 & 56.06 & 54.92 & 54.49 & 47.63 & 49.59 & 52.59 & 39.56 \\
~ & Ours & \bf{43.05} & \bf{44.39} & \bf{53.18} & \bf{43.07} & \bf{39.00} & \bf{53.37} & \bf{54.21} & \bf{51.57} & \bf{43.64} & \bf{41.02} & \bf{42.27} & \bf{34.28} \\
\hline
\end{tabular}
}
\end{center}
\end{table*}

\subsubsection{Analysis on Hyper-parameters}
The hyper-parameters $\lambda$ and $\mu$ control the quality of the adversarial training. To explore the effects of different weighting factors on defense performance, we make comparison experiments with 32-bits hashing model, as illustrated in Fig. \ref{fig:ablation}. For $\lambda$, as shown in Fig. \ref{fig:ablation:lambda}, when $\lambda$ increases, the defense performance increases until $\lambda = 1$, but the MAP values of original samples drop, which indicates there is a trade-off between robustness and precision. For $\mu$, we can observe from Fig. \ref{fig:ablation:mu} that it has a small but non-negligible impact on the outcomes, and the best behavior is achieved at $\mu=0.001$.

\subsubsection{Universality on different hashing models}
We argue that our proposed attack and defense algorithms are generic to the most popular hashing models with different backbones. To verify this point, we carry out non-targeted attacks (HAG and SDHA) on other hashing methods, including DPSH \cite{li2016feature}, HashNet \cite{cao2017hashnet} and CSQ \cite{yuan2020central}. The results are summarized in Table \ref{tab:univer}. First, for the attack, our method is still effective and better than HAG and SDHA in all cases, as shown in `No Defense' part. Moreover, even with different hashing methods or backbone networks, our defense method can still effectively mitigate the impact of adversarial attacks. Furthermore, when testing with original samples (`Original' in Table \ref{tab:univer}), or under the attacks of HAG and SDHA, the results of SAAT are higher than ATRDH, which shows that hashing models trained by our SAAT are more robust than ATRDH. Hence, the above phenomena confirm the universality of the proposed attack and defense methods.

\section{Conclusion} \label{sec:conclusion}
In this paper, we proposed a generic Semantic-Aware Adversarial Training (SAAT) algorithm for deep hashing-based retrieval, which is the first work to formulate adversarial training of deep hashing as a unified minimax problem. Specifically, we provided a discriminative mainstay features learning strategy to obtain high-quality mainstay codes as the optimal semantic representatives of samples for adversarial learning in deep hashing. Moreover, we took the mainstay code as `label' to guide the adversarial attack, where the Hamming distance between the mainstay code and the hash code of adversarial example was maximized. Furthermore, benefiting from the high-quality mainstay features, we developed a well-conceived adversarial training scheme based on a formalized minimax optimization paradigm. Extensive experiments demonstrated that our method could attain state-of-the-art results in both adversarial attack and defense on deep hashing-based retrieval.



\bibliographystyle{IEEEtran}
\bibliography{final_version}


\begin{IEEEbiography}[{\includegraphics[width=1in,height=1.25in,clip,keepaspectratio]{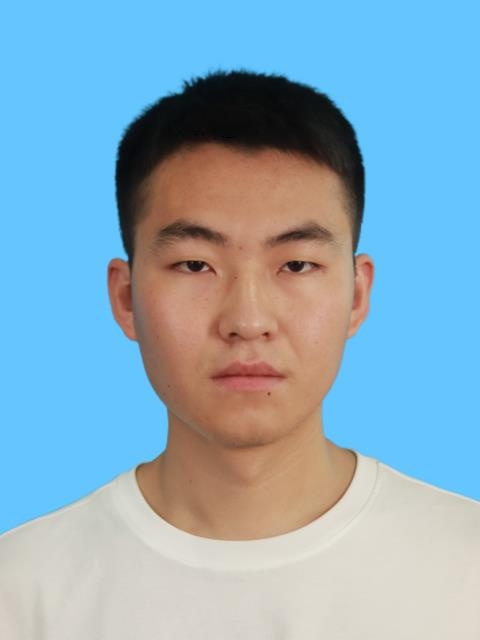}}]{Xu Yuan}
received the B.S. degree from the Harbin Institute of Technology, Weihai, China in 2021. He is currently a master student at Harbin Institute of Technology, Shenzhen, China. His research interests include deep learning and adversarial machine learning.
\end{IEEEbiography}

\begin{IEEEbiography}[{\includegraphics[width=1in,height=1.25in,clip,keepaspectratio]{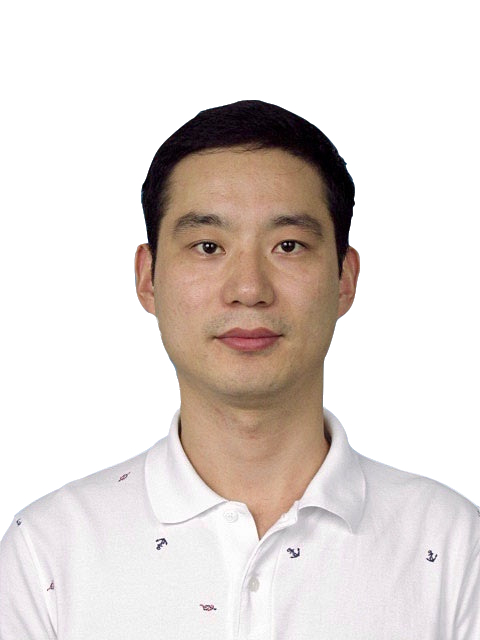}}]{Zheng Zhang} (IEEE senior member)
received Ph.D. degree in Computer Applied Technology from Harbin Institute of Technology, China. He was a Postdoctoral Research Fellow at The University of Queensland, Australia. He is currently with Harbin Institute of Technology, Shenzhen, China. He has published over 100 peer-reviewed papers at prestigious international journal and conference venues. He serves as an Associate Editor of IEEE Trans. on Affective Computing (TAFFC) and IEEE Journal of Biomedical and Health Informatics (J-BHI). His current research interests include multimedia content analysis and understanding. 
\end{IEEEbiography}

\begin{IEEEbiography}[{\includegraphics[width=1in,height=1.25in,clip,keepaspectratio]{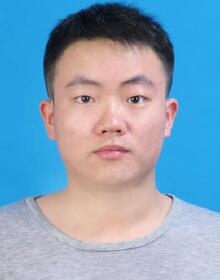}}]{Xunguang Wang}
received the B.S. degree from the China University of Geosciences, Wuhan, in 2019. He received the M.S. degree from Harbin Institute of Technology, Shenzhen, in 2022. His research interests include deep learning and adversarial machine learning.
\end{IEEEbiography}

\begin{IEEEbiography}[{\includegraphics[width=1in,height=1.25in,clip,keepaspectratio]{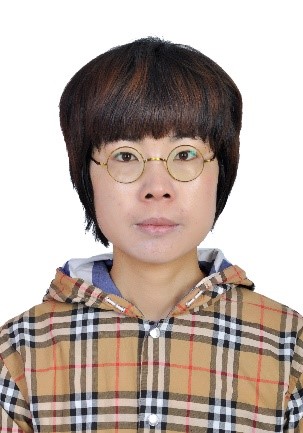}}]{Lin Wu} (IEEE senior member)
received her PhD from The University of New South Wales, Kensington, Sydney, Australia. She has broad research interests across computer vision tasks and machine learning with strong motivation to address real-world challenges and has intensively published more than 70 peer-reviewed research articles (including two book chapters) in premier journals and proceedings. Dr Wu is specialized in video content understanding (such as video instance/panoptic/object segmentation, object detection/tracking), target re-identification, incremental learning, Generative AI, and AI-based advancement for medical study. She won the 2nd Place Award in the Pixel-pixel Video Understanding in the Wild Challenge VPS Track, IEEE/CVF Computer Vision and Pattern Recognition (CVPR) 2023. She was a recipient of the Award for Growth of 2021 The 4th Eureka International Innovation and Entrepreneurship Competition (Eureka IIEC 2021, Melbourne, Australia). Dr Wu is currently serving as the Associate Editor with IEEE Trans on Neural Networks and Learning Systems (IF: 14.255), IEEE Trans on Multimedia (IF: 8.182), IEEE Trans on Big Data (IF: 4.271), and Pattern Recognition Letters (IF: 4.757). She was invited to serve as Area Chair with ACM Multimedia 2023.
 
\end{IEEEbiography}

\vfill
\end{document}